\documentclass[10pt,twocolumn,letterpaper]{article}

\usepackage{iccv}
\usepackage{times}
\usepackage{epsfig}
\usepackage{graphicx}
\usepackage{amsmath}
\usepackage{amssymb}
\usepackage{subcaption}

\usepackage{enumitem}
\usepackage{multirow}
\usepackage{flushend}
\usepackage{soul}
\newtheorem{theorem}{Theorem}
\usepackage{placeins}
\usepackage{marginnote}

\newcommand{\colimage}[3]{\begin{subfigure}{#1\textwidth}\centering\includegraphics[width=\linewidth]{#2} #3 \end{subfigure}}

\def\mypar#1{{\noindent\bf #1.}}
\def\sect#1{Section~\ref{sec:#1}}

\newcommand{\fig}[1]{Figure~\ref{fig:#1}}
\newcommand{\tab}[1]{Table~\ref{tab:#1}}

\newcommand{\pageimage}[3]{
 \begin{figure*}[h!]
  \centering
 \colimage{#1}{figs/voro32/#2mc.jpeg}{}
 \colimage{#1}{figs/voro32/#2dc.jpeg}{}
 \colimage{#1}{figs/voro32/#2sap.jpeg}{}
 \colimage{#1}{figs/voro32/#2ours.jpeg}{}
 \colimage{#1}{figs/vorogt/#2gt.jpeg}{}
  \vskip 0pt
 \colimage{#1}{figs/voro64/#2mc.jpeg}{}
 \colimage{#1}{figs/voro64/#2dc.jpeg}{}
 \colimage{#1}{figs/voro64/#2sap.jpeg}{}
 \colimage{#1}{figs/voro64/#2ours.jpeg}{}
 \colimage{#1}{figs/vorogt/#2gt.jpeg}{}
  \vskip 0pt
 \colimage{#1}{figs/voro128/#2mc.jpeg}{\caption{MC~\cite{lorensen1987marching}}}
 \colimage{#1}{figs/voro128/#2dc.jpeg}{\caption{DC~\cite{ju2002dual}}}
 \colimage{#1}{figs/voro128/#2sap.jpeg}{\caption{SAP~\cite{peng2021shape}}}
 \colimage{#1}{figs/voro128/#2ours.jpeg}{\caption{Ours}}
 \colimage{#1}{figs/vorogt/#2gt.jpeg}{\caption{Ground Truth}} 
   \caption{Visual comparison of optimization-based methods for grids of size $32^3$ (top row), $64^3$ (middle row), and $128^3$ (bottom row)}\label{#3}
 \end{figure*}}

\newcommand{\abcpageimage}[4]{
 \begin{figure*}[h!]
  \centering
 \colimage{#1}{figs/renders_ABC32/#2NMC.jpeg}{}
 \colimage{#1}{figs/renders_ABC32/#2NDC.jpeg}{}
 \colimage{#1}{figs/renders_ABC32/#2ours.jpeg}{}
 \colimage{#1}{figs/renders_ABC32/#2gt.jpeg}{}
  \vskip 0pt
    \centering
 \colimage{#1}{figs/renders_ABC64/#2NMC.jpeg}{}
 \colimage{#1}{figs/renders_ABC64/#2NDC.jpeg}{}
 \colimage{#1}{figs/renders_ABC64/#2ours.jpeg}{}
 \colimage{#1}{figs/renders_ABC64/#2gt.jpeg}{}
  \centering

  \centering
 \colimage{#1}{figs/renders_ABC32/#3NMC.jpeg}{}
 \colimage{#1}{figs/renders_ABC32/#3NDC.jpeg}{}
 \colimage{#1}{figs/renders_ABC32/#3ours.jpeg}{}
 \colimage{#1}{figs/renders_ABC32/#3gt.jpeg}{}
  \vskip 0pt
    \centering
 \colimage{#1}{figs/renders_ABC64/#3NMC.jpeg}{\caption{NMC~\cite{chen21tog}}}
 \colimage{#1}{figs/renders_ABC64/#3NDC.jpeg}{\caption{NDC~\cite{chen22tog}}}
 \colimage{#1}{figs/renders_ABC64/#3ours.jpeg}{\caption{Ours}}
 \colimage{#1}{figs/renders_ABC64/#3gt.jpeg}{\caption{Ground Truth}}
   \caption{Visual comparison of learning-based methods for grids of size $32^3$ (top row), $64^3$ (bottom row)}\label{#4}
 \end{figure*}}

\newcommand{\thingipageimage}[3]{
 \begin{figure*}[h!]
  \centering
 \colimage{#1}{figs/voro32/#2NMC.jpeg}{}
 \colimage{#1}{figs/voro32/#2NDC.jpeg}{}
 \colimage{#1}{figs/voro32/#2oursRC.jpeg}{}
 \colimage{#1}{figs/vorogt/#2gt.jpeg}{}
  \vskip 0pt
\colimage{#1}{figs/voro64/#2NMC.jpeg}{}
 \colimage{#1}{figs/voro64/#2NDC.jpeg}{}
 \colimage{#1}{figs/voro64/#2oursRC.jpeg}{}
 \colimage{#1}{figs/vorogt/#2gt.jpeg}{}
  \vskip 0pt
 \colimage{#1}{figs/voro128/#2NMC.jpeg}{\caption{NMC~\cite{chen21tog}}}
 \colimage{#1}{figs/voro128/#2NDC.jpeg}{\caption{NDC~\cite{chen22tog}}}
 \colimage{#1}{figs/voro128/#2oursRC.jpeg}{\caption{Ours}}
 \colimage{#1}{figs/vorogt/#2gt.jpeg}{\caption{Ground Truth}}
   \caption{Visual comparison of learning-based methods for grids of size $32^3$ (top row), $64^3$ (middle row), and $128^3$ (bottom row)}\label{#3}
 \end{figure*}}

\makeatletter
\DeclareRobustCommand\onedot{\futurelet\@let@token\@onedot}
\def\@onedot{\ifx\@let@token.\else.\null\fi\xspace}

\def\etal{\emph{et al}\onedot}
\makeatother

\usepackage[pagebackref=true,breaklinks=true,letterpaper=true,colorlinks,bookmarks=false]{hyperref}

\iccvfinalcopy 

\def\iccvPaperID{10980} 

\ificcvfinal  \fi

\begin{document}

\title{\vspace{-5mm}VoroMesh: Learning Watertight Surface Meshes with Voronoi Diagrams}

\author{
Nissim Maruani\footnotemark[1]\\[-0.5mm]
\small{Inria, Universit\'e C\^ote d'Azur}\\[-1mm]
{\tt\small nissim.maruani@inria.fr}
\and
Roman Klokov\footnotemark[1] \\[-.5mm]
\small{LIX, École Polytechnique, IP Paris}\\[-1mm]
{\tt\small klokov@lix.polytechnique.fr}
\and
Maks Ovsjanikov\\[-.5mm]
\small{LIX, École Polytechnique, IP Paris}\\[-1mm]
{\tt\small maks@lix.polytechnique.fr}
\and
Pierre Alliez\\[-.5mm]
\small{Inria, Universit\'e C\^ote d'Azur}\\[-1mm]
{\tt\small pierre.alliez@inria.fr}
\and
Mathieu Desbrun\\[-.5mm]
\small{Inria Saclay - Ecole Polytechnique}\\[-1mm]
{\tt\small mathieu.desbrun@inria.fr}
}
\maketitle
\ificcvfinal \thispagestyle{empty} \fi
\renewcommand{\thefootnote}{\fnsymbol{footnote}}
\footnotetext[1]{Authors' roles and contributions are listed in Section~\ref{sec:Contributions}}
\begin{abstract}
In stark contrast to the case of images, finding a concise, learnable discrete representation of 3D surfaces remains a challenge. In particular, while polygon meshes are arguably the most common surface representation used in geometry processing, their irregular and combinatorial structure often make them unsuitable for learning-based applications.
In this work, we present VoroMesh, a novel and differentiable \emph{Voronoi-based representation} of watertight 3D shape surfaces. 
From a set of 3D points (called generators) and their associated occupancy, we define our boundary representation through the Voronoi diagram of the generators as the subset of Voronoi faces whose two associated (equidistant) generators are of opposite occupancy: the resulting polygon mesh forms a watertight approximation of the target shape's boundary.
To learn the position of the generators, we propose a novel loss function, dubbed \emph{VoroLoss}, that minimizes the distance from ground truth surface samples to the closest faces of the Voronoi diagram 
which does not require
an explicit construction of the entire Voronoi diagram. 
A direct optimization of the Voroloss to obtain  generators on the Thingi32 dataset  demonstrates the geometric efficiency of our representation compared to axiomatic meshing algorithms and recent learning-based mesh representations. 
We further use VoroMesh in a learning-based mesh prediction task from input SDF grids on the ABC dataset, and show comparable performance  to state-of-the-art methods while guaranteeing closed output surfaces free of self-intersections.
\end{abstract}

\begin{figure}[h]
 \centering
 \begin{subfigure}{.15\textwidth}
  \centering
  \includegraphics[width=\linewidth]{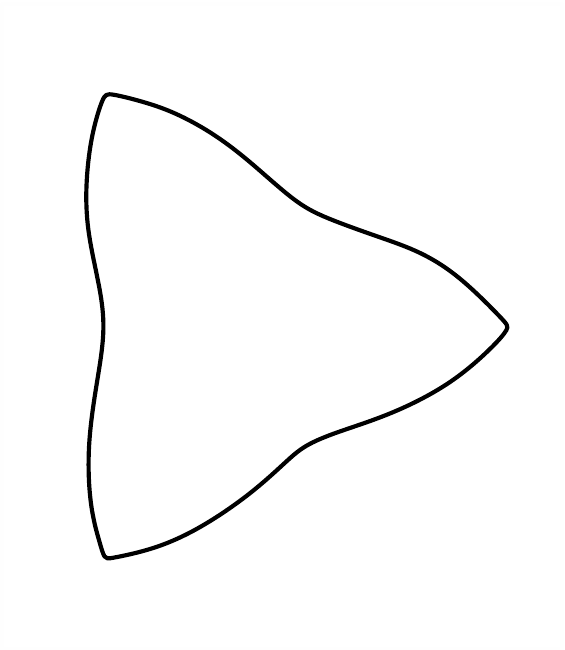}  
\end{subfigure}
\begin{subfigure}{.15\textwidth}
  \centering    
   \includegraphics[width=\linewidth]{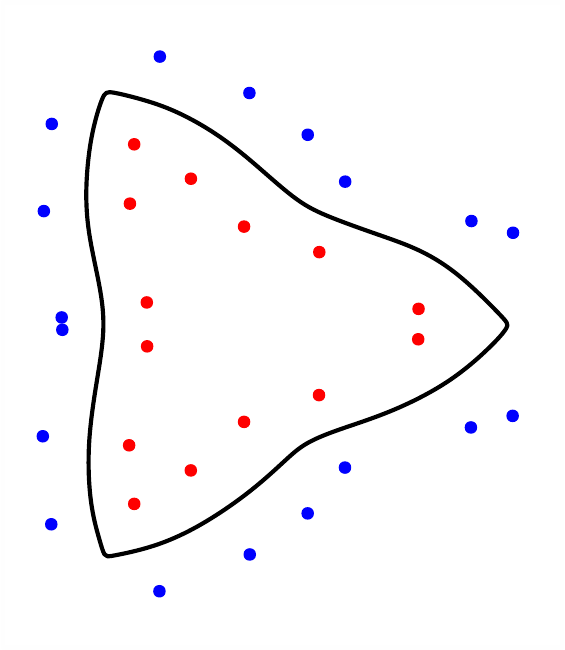}  
\end{subfigure}
\begin{subfigure}{.15\textwidth}
  \centering
   \includegraphics[width=\linewidth]{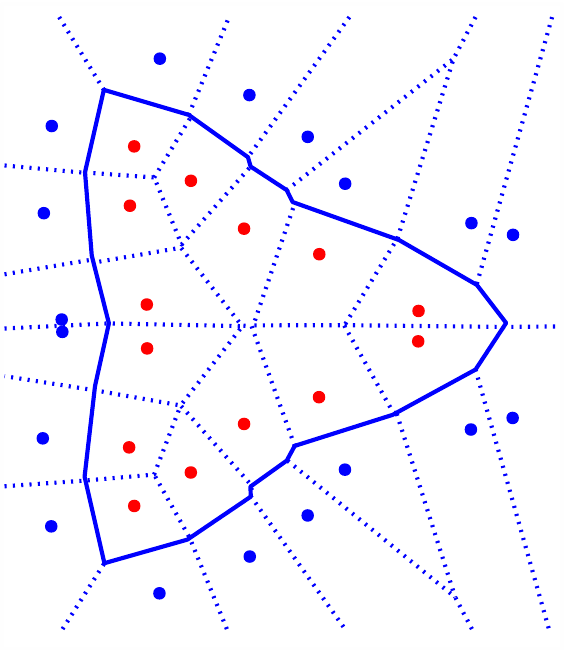}  
\end{subfigure}

 \begin{subfigure}{.155\textwidth}
  \centering
  \includegraphics[width=\linewidth]{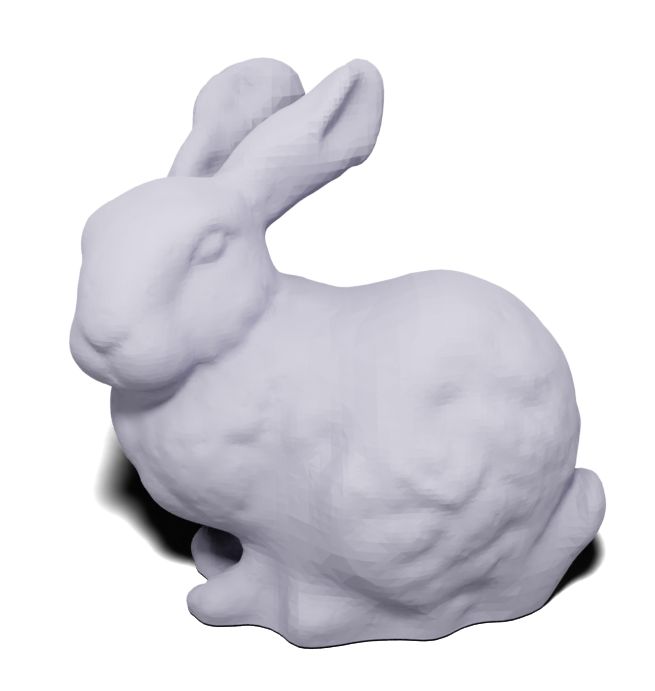} 
    \caption{Target surface}
\end{subfigure}
\begin{subfigure}{.155\textwidth}
  \centering    
  \includegraphics[width=\linewidth]{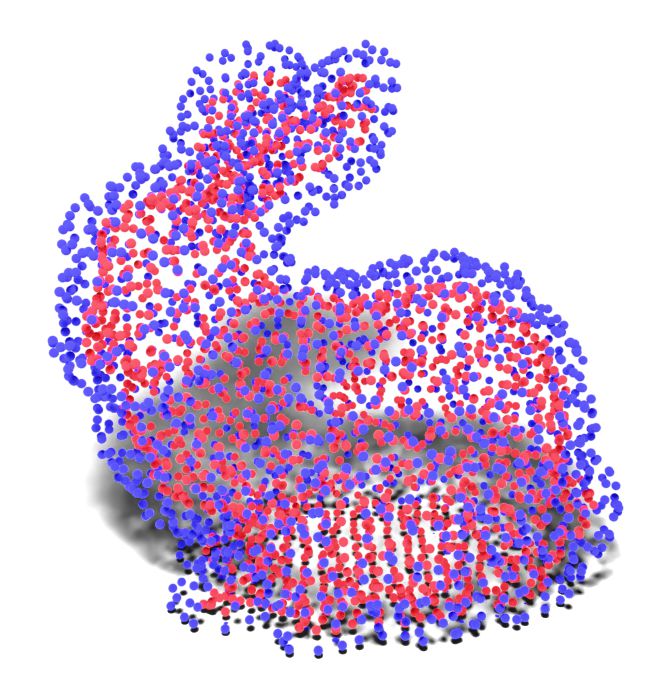}
\caption{Optimization}
\end{subfigure}
\begin{subfigure}{.155\textwidth}
  \centering
  \includegraphics[width=\linewidth]{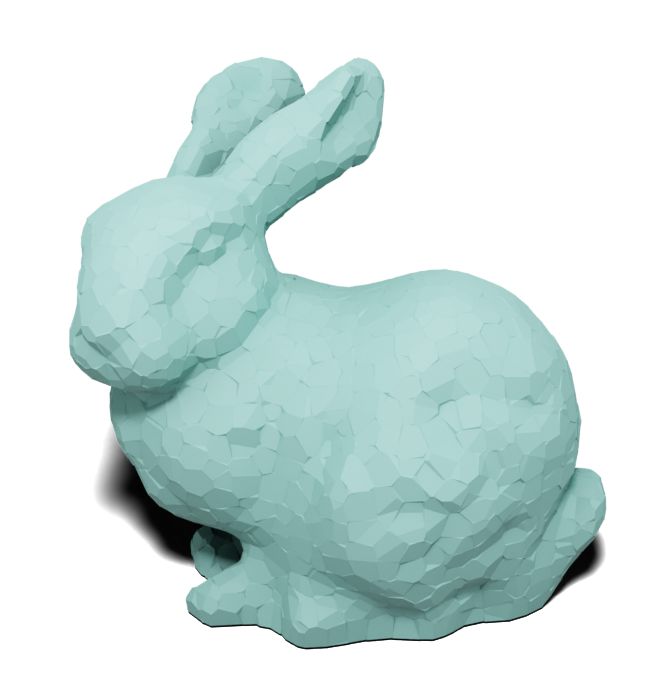}  
\caption{\textit{VoroMesh}}
\end{subfigure} \vspace*{-3mm}
\caption{\textbf{VoroMesh at a glance}: To fit our representation to (a) a target shape (top: curve; bottom: surface), (b) the positions of generators around the shape boundary are optimized based on our \textit{VoroLoss}, before a binary occupancy is assigned to each generator; (c) the \textit{VoroMesh} is then extracted as a subset of the Voronoi diagram of the generators containing only the Voronoi faces between pairs of generators of opposite occupancy. \vspace*{-4mm}
  }
  \label{fig:summary}
\end{figure}

\section{Introduction}
\label{sec:intro}

Geometry processing of three-dimensional shapes typically relies on a discrete representation of its surface boundary. Among various boundary representations, point sets and surface polygon meshes are popular because they are simple to generate, edit, or render. 
While most tasks can be efficiently performed with non-learning tools from the standard geometry processing toolbox, some of them (e.g., ill-posed problems such as reconstruction from sparse data, or data-driven tasks) require learning tools. Using neural networks with a differentiable (explicit or implicit) boundary representation can be a powerful way to tackle these problems. 
Yet, discrete representations such as surface polygon meshes rely on two parts: vertex locations, and vertex connectivity defining the polygonal facets. The first part, continuous in nature, is easy to learn, but the second part, now combinatorial, is substantially more difficult, which has hampered mesh-based learning methods in the past. Indeed, 
besides the complexity issues arising from modeling vertex connectivity, the discrete nature of topological configurations in meshes poses two accompanying differentiability issues that are particularly relevant in learning applications.
First, the way connectivity is induced from corresponding discrete variables generally is not differentiable; even if probabilistic modeling can offer differentiable sampling from discrete variables~\cite{jang17iclr}, the combinatorial nature of connectivity forces faces and edges to be considered independently, which often produce polygon soup predictions instead of watertight meshes.
Second, vertex position optimization can also be hindered since non-stable mesh prediction during training results in discontinuity in most of the commonly-used loss functions based on sampling from predicted mesh surfaces, such as chamfer distance and point-to-surface distance: 
any change of topology in the prediction over the course of training can result in discontinuous changes in the coordinates of on-surface samples.

State-of-the-art data-driven methods for mesh prediction can be roughly categorized into two groups: methods producing shapes with neural implicit representations and meshing their zero-level sets~\cite{guillard22eccv,michalkiewicz19iccv,park19cvpr,remelli20neurips}, and approaches predicting explicit meshes as sets of vertices and faces~\cite{chen22tog,chen21tog,gao20neurips,nash20icml,shen21neurips,wang18eccv}. While proven to be powerful at capturing surfaces, implicit representations have their drawbacks, such as the need for additional test-time surface discretization to produce meshes (which can be costly for high resolutions), and limited control over the geometry of the produced meshes due to the inability to explicitly define optimization objectives on the output surfaces. While recent works in the second category show promising results in explicit mesh prediction, most of them rely on the use of regular discretization grids, which limit their expressivity and do not guarantee watertightness of the output meshes. 

In order to circumvent the issues mentioned thus far, we propose to use a differentiable surface representation based on the \emph{3D Voronoi diagram}~\cite{aurenhammer1991voronoi} of an input point set of generators which canonically derives its combinatorial information from their location. The continuous placement of generators results in an adaptive discretization of 3D surfaces which substantially improves the expressivity of our representation compared to methods relying on regular grids.
We leverage the geometric properties of the Voronoi diagram (specifically, the fact that Voronoi cells are defined as intersections of halfspaces) to formulate a novel loss based only on bisectors between generators rather than the full Voronoi diagram: this allows for a differentiable optimization of the 3D Voronoi generators. Since our loss function only requires the position of generators for faithful surface approximation, our Voronoi representation successfully mitigates the differentiability issues associated with explicit connectivity modeling. Compared to previous Voronoi-based representations~\cite{williams2020voronoinet}, we do not focus on volume decomposition but on surface approximation, so our loss does not need to be regularized via the addition of the centroidal Voronoi tessellation energy to favor the spreading of generators. \smallskip

\noindent In summary, the main contributions of this work are:
\begin{itemize}[leftmargin=3mm,itemsep=-0.5ex,labelsep=0.6ex,topsep=0.3ex]
    \item We establish a novel loss function, the \textit{VoroLoss}, which directly optimizes 3D Voronoi generators to fit an input target shape (provided as a surface triangle mesh) without requiring an explicit computation of the Voronoi diagram;
    \item We show that the representation power of our \textit{VoroMesh} outperforms Marching Cubes, Dual Contouring, and two recent learning approaches in terms of geometric fidelity;
    \item We demonstrate that the proposed \textit{VoroMesh} and \textit{VoroLoss} can be integrated into a learning pipeline to predict 3D Voronoi generators, along with their occupancy, to reconstruct closed and non-self-intersecting meshes from input regular grids of SDF values.
\end{itemize}

\section{Related Work}
\label{sec:related}

Numerous prior works have sought 3D shape representations amenable to learning. We review the most relevant approaches while pointing the reader to a recent thorough review on intelligent mesh generation~\cite{li2022survey} --- which, notably, identifies the generation of watertight manifold meshes as one of the most significant open challenges.\smallskip

\mypar{Implicit representations}
To bypass the issues of shape discretization, pioneering 3D learning methods proposed to use implicit surface representations defining shape boundaries as iso-surfaces of neural distance fields~\cite{erler20eccv,genova20cvpr,park19cvpr} or as decision boundaries of neural occupancy classifiers~\cite{chen19cvpr,ma22cvpr,mescheder2019occupancy,yan22cvpr}. Most of the earlier methods, however, rely on global features, trying to encode complete surface information into a single global latent vector. This constraint limits the ability of these models to encode sharp local features and prevents efficient generalization of models across different categories and data domains. Additionally, most of these works resort to Marching Cubes~\cite{lorensen1987marching} for the final mesh extraction, needing a cubic number of forward passes of the occupancy/distance prediction network. 
As a consequence, sharp features partially captured by the implicit function are further smoothed out at low resolutions. 
Single shape optimization~\cite{gropp20icml,sitzmann2020implicit} is yet another type of implicit representation, for which a shape is encoded by fitting a whole neural network; but learning from previous data is not possible, preventing many data-driven applications. 
Although our work focuses on the prediction of explicit discrete surface representation, it could potentially be combined similarly to related mesh prediction approaches discussed below, since inferred SDF values could be used as inputs of our model.\smallskip

\mypar{Discrete representations}
Different shape representations have been proposed to capture geometric information for learning-based applications.
Early 3D shape modeling methods used regular~\cite{klokov19bmvc,wu16neurips} and adaptive~\cite{tatarchenko17iccv} grids of binary occupancy, or their decomposition~\cite{richter18cvpr} predicted by 3D CNNs to represent reconstructed geometry. 
A separate line of works proposed to predict shapes as sets of points inferred with MLPs~\cite{achlioptas18icml,su17cvpr}, invertible normalizing flows~\cite{klokov20eccv,yang19iccv}, and diffusion-based generative models~\cite{luo21cvpr,zeng22neurips}. Our representation relates to point clouds in two aspects: we predict continuous point coordinates to represent shapes and use distance-based loss functions to provide the training signal; but our representation does not restrict output points to lie on the surface, using them instead as generators of a 3D Voronoi diagram fitting the surface.

Occupancy grids and point clouds can be viewed as approximations of target surfaces suitable for prediction with neural networks and continuous optimization, but containing only partial mesh information. Despite the challenges stated in Section \ref{sec:intro}, numerous works have proposed models capable of producing meshes. Deformation-based approaches~\cite{tan18cvpr} use template meshes and predict vertex offsets to fit target surfaces without any changes in the initial connectivity. The approach of Nash \etal~\cite{nash20icml} treats vertices and connectivity as random variables and uses an autoregressive probabilistic model to derive likelihood-based objectives, suitable for optimization. Several works also proposed to model surfaces with predicted parametric primitives~\cite{chen20cvpr,deng20cvpr}, allowing differentiation by continuous relaxation of discrete variables. However, representations from this family do not generalize well to new data. 

Most relevant to our work are methods that lift the limitation of global shape encoding and use local features obtained from coarse, noisy, or partial inputs. 
Deformable volumetric tetrahedral meshes~\cite{gao20neurips,shen21neurips} for instance can synthesize shapes from SDF grids. While achieving good performance, these approaches mainly compare themselves to global models, do not provide topological guarantees on the output shapes, and suffer from artifacts due to their use of regular grids.  
Deep Marching Cubes~\cite{liao18cvpr} propose differentiable point-to-mesh distance and curvature losses as expectations over the space of topological configurations to train occupancy and vertex displacements on a grid. Since occupancy is modeled independently, this approach is unable to provide guarantees on topological correctness. 
A similar approach introduces a new representation inspired by Marching Cubes to capture topological and vertex information~\cite{chen21tog} with extended sets of discrete and continuous variables for every vertex of a regular grid. Two separate networks predict these variables independently for each grid vertex, so they form a consistent output mesh. Follow-up work~\cite{chen22tog} substitutes these variables with a reliance on the mesh produced by Dual Contouring~\cite{ju2002dual} and extends the approach to various types of input data. Both of these approaches circumvent the necessity to differentiate through meshing procedures by the construction of ground truth proxy data structures during pre-processing and training networks to infer them from input data. However, resulting meshes are not always manifold for NDC~\cite{chen22tog}, and both methods often produce self-intersections. 
Shape-as-points~\cite{peng2021shape} predicts coordinate offsets and normals for noisy point cloud inputs which are used to produce indicator grids with a differentiable version of PSR~\cite{kazhdan06esgp}. These indicator grids are supervised with precomputed grids obtained from ground truth surface samples. 
Compared to these works, 1) we do not construct output meshes during training, thus avoiding the pitfalls associated with differentiation through explicit meshing; 2) supervision relies on sampled points and does not require any pre-computed data structures; 3) our approach is not limited in expressivity by the use of a fixed grid, since it allows adaptive placement of generators.\\[-3.4mm]




\mypar{Shape representation via computational geometry}
A few works have already exploited Voronoi diagrams to represent 3D shapes \cite{abdelkader2020vorocrust, amenta2001power}. However, providing a differentiable loss to compute such a diagram is not straightforward. VoronoiNet \cite{williams2020voronoinet}, for instance, uses a soft definition of the Voronoi diagram to match the occupancy of a given array in 2D applications. However, their formulation offers a differentiable implicit representation of \emph{solid objects}, in the sense that they spread generators across the whole volume 
(through a CVT-based regularizer) to offer a convex decomposition of the inside and outside of the 2D object, thus extending CvxNet~\cite{deng20cvpr}. (A single 3D example, a sphere, was exhibited.) Instead, we focus on  \emph{approximating the boundary of a 3D object}: as a consequence, far fewer generators are required and they remain close to the boundary as their role is purely to approximate the surface well, capturing the normal field and sharp features of the surface. Our introduction of a dedicated loss also removes the need to use a soft-argmin version of Voronoi diagrams. Finally, other related works have proposed to use the dual of Voronoi diagrams, i.e., Delaunay triangulations, to offer a differentiable approach to surface triangulations~\cite{rakotosaona2021differentiable,rakotosaona2021learning}; however, they do not offer topological guarantees due typically to the presence of ``non-manifold'' triangles in resulting meshes.

\section{Voronoi Mesh Representation}
\label{sec:method}

We now describe the VoroMesh approach by first explaining our Voronoi-based representation via generators, then the \textit{VoroLoss} which minimizes the distance from an input point cloud sampling a surface to Voronoi facets of the generators, then occupancy determination and mesh extraction. Finally, we discuss the optimization of this representation for a given input shape and its use in learning tasks.

\subsection{VoroMesh Overview}
Our VoroMesh representation canonically defines a watertight mesh via a Voronoi diagram. Given a set of $N$ distinct 3D points that we call generators $\textbf{Q} \!=\! \{q_i \!\in\! \mathbb{R}^3\} $, the \emph{Voronoi cell} $V_i$ of a generator $q_i$ is the set of points $y \!\in\! \mathbb{R}^{3}$ whose distance to $q_i$ is smaller than their distance to any other site $q_j$. Each $V_i$ is known to be a convex polytope, whose boundary is composed of flat polygonal \emph{Voronoi faces} lying on bisectors between $q_i$ and some of its closest neighboring generators $q_j$. The union of these disjoint 3D regions forms a convex decomposition of $\mathbb{R}^{3}$~\cite{aurenhammer1991voronoi}. We assign a binary occupancy $\textbf{O} \!=\! \{o_i\} \!\in\! \{0,1\}^N$ to each generator, such that the union of the cells with non-zero occupancy yields a 3D volume composed of convex polyhedral cells. The boundary of this 3D volume is then a Voronoi-induced surface mesh, that we call $\textit{VoroMesh}(\textbf{Q}, \textbf{O})$. Assuming that all infinite cells are tagged as outside, we obtain a \emph{watertight surface mesh} (i.e., without holes or self-interpenetrations, and with a clearly defined inside). Our approach then consists in optimizing the generators so that induced Voronoi faces best fit the 3D ground truth shape.\smallskip

\mypar{VoroLoss loss function}
For optimization and learning purposes, we need to be able to minimize the distance between a VoroMesh and a dense sampling $X$ of a target shape. 
Computing the entire Voronoi diagram induced by the generators at each step of the optimization would be too costly. Instead, we leverage the geometric properties of Voronoi diagrams (and in particular, their locality) to find a loss function that can directly provide the correct distance without invoking a full Voronoi diagram construction. More specifically, we optimize the generators \textbf{Q} not by minimizing the distance from $X$ to the current Voronoi faces, but by minimizing the distance from $X$ to \emph{all} the faces of the full diagram, i.e., $\sum_{x \in X} \min_i \|x \!-\! \partial V_i\|^2$ --- which can efficiently be achieved by only considering all close pairs of generators. 
This property stems from the following theorem (see supplementary material for a proof):
\begin{theorem}
 The distance from $x$ to its closest face in a Voronoi diagram equals the distance from $x$ to the closest bisector $H_{i_x, j}$ formed between $q_{i_x}$ whose Voronoi cell contains $x$ and another Voronoi site $q_j$: 
 \[ \|x - \partial V_i\| = \min_{j \neq i_x}  \|x-H_{i_x, j}\|. \]
\end{theorem}

\noindent We thus introduce a loss function, dubbed \textit{VoroLoss}:
\[\textit{VoroLoss}(X, \textbf{Q}):=\sum_{x \in X} \min_{j\neq i_x} \|x-H_{i_x, j}\|^2 \]

The simplification implied by evaluating distances to planes instead of distances to the polygonal faces of a computed Voronoi diagram allows for a simple and efficient implementation. In practice, we use a $k$-nearest neighbor algorithm to select the closest cells $i_x$ and the closest bisector planes $H_{i_x, j}$ to cull a large number of computations. For $N$ generators and a sampling $X$ of size $|X|\!=\!D$, the average complexity for computing this loss becomes $\mathcal{O}(ND)$.  Note finally that the \textit{VoroLoss} is computed without any occupancy information: it simply tries to make each sample be on a face of the Voronoi diagram of the generators.\smallskip

\mypar{Occupancy and watertight mesh extraction}
Once the \emph{VoroLoss} is minimized, we must determine the occupancy of each generator (see colors of the generators in \fig{summary}). 
This is quickly achieved by simply using the ground truth occupancy of the barycenter of each Voronoi cell in practice. 
Once the occupancy of all generators is established, the polygon mesh can be extracted by keeping the Voronoi faces separating generator pairs of complementary occupancy. 
By construction, this mesh represents the boundary of a volume: it thus has a separate interior and exterior, and it is self-intersection free. In theory, cospherical (resp. cocircular) points of different occupancies may create non-manifold vertices (resp. edges). To fix these topological exceptions
one can simply duplicate the corresponding vertices (resp. edges). Note however that in our tests, such a case never happened in practice. \smallskip


\mypar{Shape optimization}
Based on the building blocks we described thus far, we can now fit a $\textit{VoroMesh}(\textbf{Q}, \textbf{O})$ to a target surface as follows (see \fig{summary}):

\begin{enumerate}[leftmargin=4mm,itemsep=-0.5ex,labelsep=0.6ex,topsep=0.2ex]
\item Densely sample a set of points $X \!\in\!  \mathbb{R}^{M \times 3}$ from a ground truth surface mesh $(V,F)$;
\item Initialize $\textbf{Q} \in  \mathbb{R}^{N \times 3}$; 
\item Minimize $\textit{VoroLoss}(X, \textbf{Q})$;
\item Compute the Voronoi diagram of $\textbf{Q}$;
\item Determine the ground truth occupancy $\textbf{O}$ of the barycenter of each Voronoi cell;
\item Compute the final polygonal mesh $\textit{VoroMesh}(\textbf{Q}, \textbf{O})$.
\end{enumerate}
For Step 2, we initialize the generators $\textbf{Q}$ by selecting, out of a regular voxel grid, every grid node adjacent to one or more voxels containing samples from $X$.

\begin{figure}[h]
 \centering
 \begin{subfigure}{.09\textwidth}
  \centering
  \includegraphics[width=\linewidth]{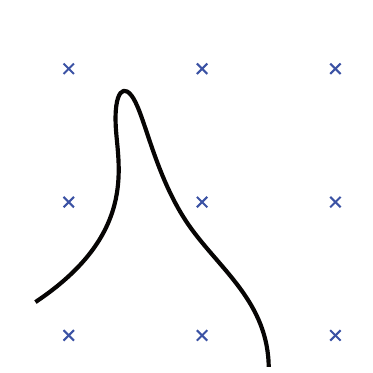}  
  \caption{Target}
\end{subfigure}
 \begin{subfigure}{.09\textwidth}
  \centering
  \includegraphics[width=\linewidth]{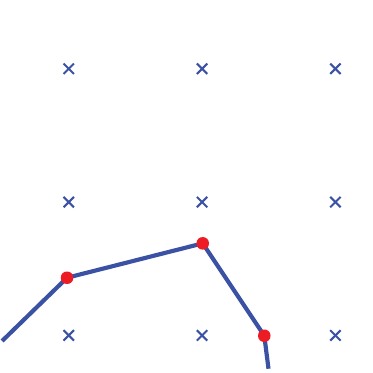}  
  \caption{MC~\cite{lorensen1987marching}}
\end{subfigure}
\begin{subfigure}{.09\textwidth}
  \centering    
   \includegraphics[width=\linewidth]{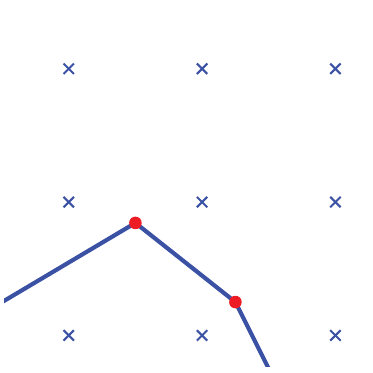}  
   \caption{DC~\cite{ju2002dual}}
\end{subfigure}
\begin{subfigure}{.09\textwidth}
  \centering
   \includegraphics[width=\linewidth]{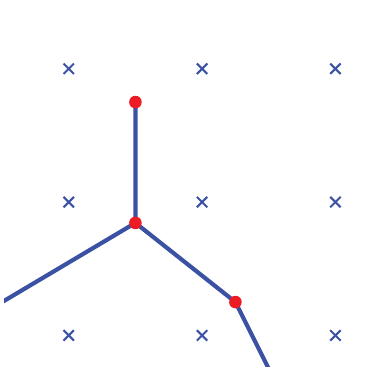}  
   \caption{UDC~\cite{chen22tog}}
\end{subfigure}
\begin{subfigure}{.09\textwidth}
  \centering
   \includegraphics[width=\linewidth]{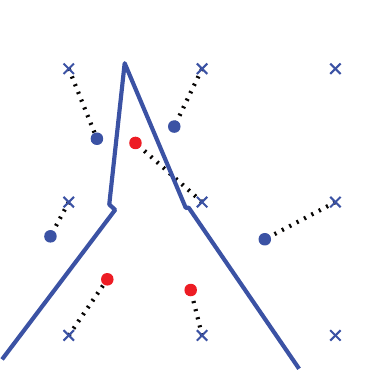}  
   \caption{Ours}
\end{subfigure}
  \caption{Marching Cubes (b) and Dual Contouring (c) cannot capture details of a target shape (a) smaller than the grid size; UDC (d), based on edge-crossings, can but at the price of a non-manifold elements. \textit{VoroMesh} (e) both captures the details and returns a closed and manifold mesh}
    \label{fig:details}
\end{figure}

\subsection{Learning VoroMeshes}
To demonstrate the effectiveness of our representation in data-driven settings, we focus on one application: shape reconstruction from a sparse grid sampling of the SDF. Unlike other 3D learning tasks (such as single view reconstruction), there is no need to encode complete shape information into a single global latent variable, so we can make use of specific features representing local configurations of particular parts of the volume rather than a single global representation: this will allow the network to generalize to new shapes that are strikingly different from the training set.

\begin{figure}[h]
  \centering
  \includegraphics[width=0.98\linewidth]{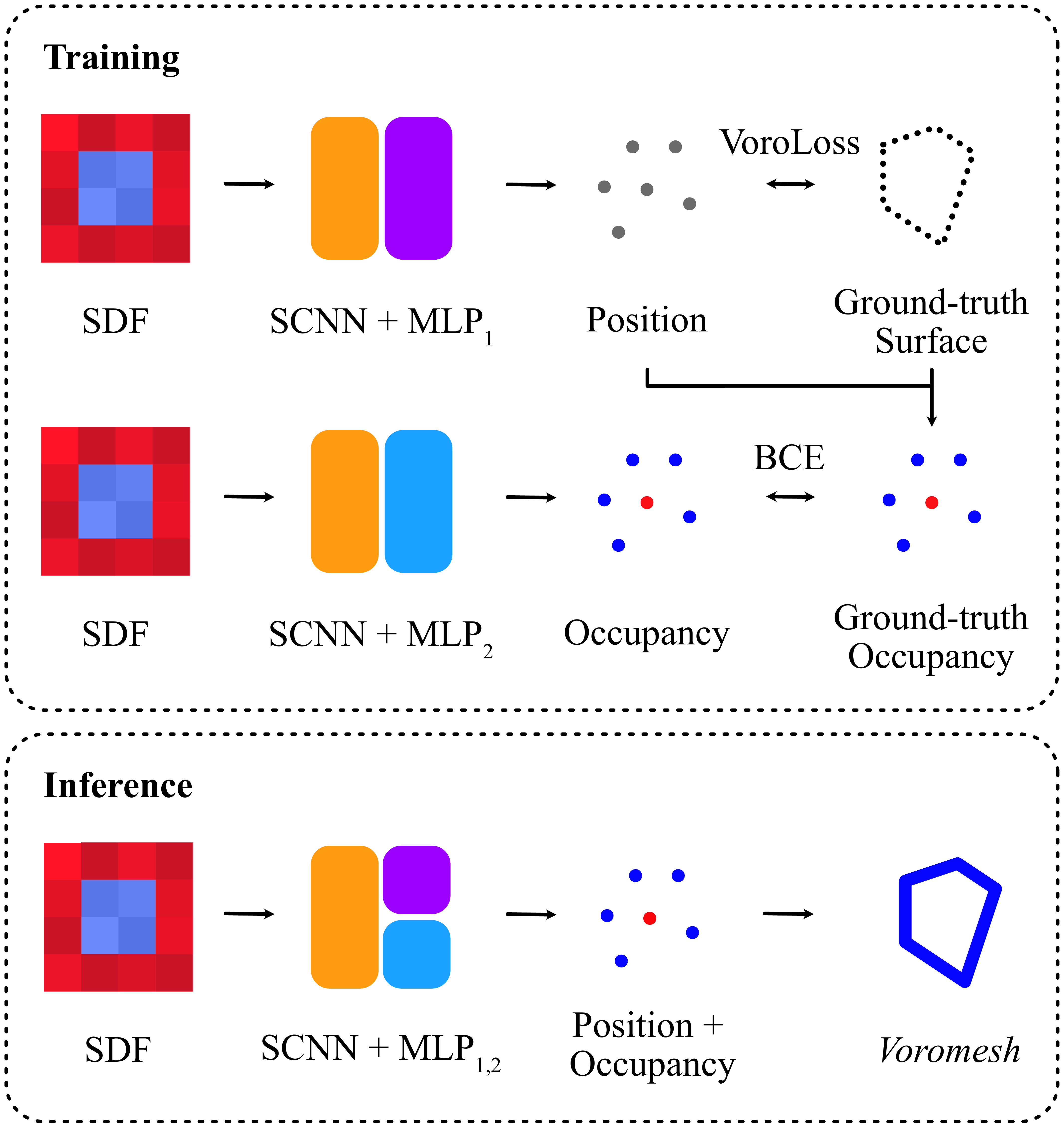}
  \caption{We train SCNN and first MLP to predict the position of generators, supervised by the \textit{VoroLoss} with respect to a sampling of the input shape (top row). We then compute the ground truth occupancy of the predicted generators, and use this information to train a second MLP (middle row). During inference, the whole network predicts the position and occupancy of each generator from which the \textit{VoroMesh} can be extracted (bottom row).\vspace*{-3mm}}
  \label{fig:summaryCNN}
\end{figure}

To generate a \textit{VoroMesh} from a grid $I \!\in\! \mathbb{R}^{N \!\times\! N \!\times\! N\!}$ of signed distances, we predict the positions of the generators along with their occupancy (see Figure~\ref{fig:summaryCNN}). Knowing that most of the information describing the surface is concentrated in the SDF values closest to the surface, we mask the SDF grid through thresholding around the isovalue and convert it into a sparse grid. Selected grid points close to the surface $v_i$ serve as the initial positions of the generators. The sparse grid of SDF values is fed to a generic ResNet-like Sparse 3D CNN (SCNN)~\cite{choy19cvpr} with a receptive field of size $19^3$ to produce local features. Using SCNNs not only alleviates the computational complexity burden of 3D convolutions, but also forces the CNN filters to focus on near-surface variations in SDF values and to spend the parameter budget efficiently, capturing information that is most relevant for surface reconstruction. We obtain the final position $q_i$ and occupancy $o_i$ of generators with two separate MLPs acting on local features produced by the SCNN, via:
\begin{equation}
  \begin{split}
    & q_i = v_i + \textrm{MLP}_1(\textrm{SCNN}(\textrm{SDF}(v_i))),\\
    & o_i = \textrm{MLP}_2(\textrm{SCNN}(\textrm{SDF}(v_i))),\\[-1mm]
  \end{split}
\end{equation}
where $\textrm{SDF}(v_i)$ is SDF values at the corresponding point $v_i$.

We optimize the parameters of our model in two steps. First, we train the SCNN and MLP$_1$ to predict the position of the generators $Q$ using our \textit{VoroLoss}. Previous offset prediction based networks often introduced hyperbolic tangent non-linearity at the end of the network to bound the values of possible offsets. We observed that at the start of optimization, this may lead to offsets quickly reaching these bounds and never returning to lower values because of the vanishing gradients of the hyperbolic tangent function. Since we do not necessarily want to force generators to stay in their initial cells and wish to obtain more informative gradients during training, we refrained from using hyperbolic tangent and instead, introduced a mild regularization on the maximum values of predicted offsets with:\vspace*{-1.5mm}
\begin{equation}
  R(\textbf{Q}) = \textrm{max}_i \, ||\Delta v_i||_2\vspace*{-1mm}
\end{equation}
which we add to \textit{VoroLoss} to form a loss $L(X, \textbf{Q})$ as:\vspace*{-1mm}
\begin{equation}
  L(X, \textbf{Q}) = \textit{VoroLoss}(X, \textbf{Q}) + \lambda R(\textbf{Q}),\vspace*{-1mm}
\end{equation}
with $\lambda$ being a regularization weight.
Unlike NDC~\cite{chen22tog}, we do not pre-compute the position of generators, but rely on our loss to move the generators in place without any mesh supervision. By not requiring mesh construction and point sampling to fit surfaces at all, our approach completely evades the difficulties typically arising from the differentiation of these operations.

The second training step freezes the convolutional part and optimizes the parameters of MLP$_2$ with minimization of the binary cross-entropy (BCE) between predicted and ground truth occupancy probabilities, extracted as discussed in the previous section. At inference time, a wrong occupancy prediction for a Voronoi cell could create an undesired surface artifact. Two fallback solutions based on queries to input SDF values can be used to fix incorrect occupancies: (1) increase the thickness of the SDF mask to include additional fixed Voronoi generators with deterministic occupancy, which will then bound the spatial scale of possible artifacts; (2) assign the occupancy of some cells using their vertices' input SDF values and check that no voxel having a vertex with signed distance larger than the diagonal length of a voxel 
does not contain the target surface. Since predicted occupancy is only needed for near-surface cells, we found that these two simple actions remove all potential issues in practice. \smallskip

A last, yet important technical remark is that compared to NMC and NDC, our implementation allows multi-shape batching and parallelized forward and backward computations during training and inference, since MLPs and SCNNs~\cite{choy19cvpr} can process several samples simultaneously.

\section{Experimental Results}
\label{sec:experiments}

\begin{table*}[t]
  \begin{center}
    \resizebox{1.\linewidth}{!}{

\begin{tabular}{|l|r|r|r|r|r|r|r|r|r|r|r|r|}
  \hline
  \multirow{2}{*}{Grid Size} & \multicolumn{3}{|c|}{CD ($\times 10^{-5}$)}  & \multicolumn{3}{c|}{F1 ($\delta = 0.003$) } & \multicolumn{3}{c}{NC} & \multicolumn{3}{|c|}{Time (s)} \\
  \cline{2-13}
   & $32^3$ & $64^3$ & $128^3$ 
   & $32^3$ & $64^3$ & $128^3$
   & $32^3$ & $64^3$ & $128^3$
   & $32^3$ & $64^3$ & $128^3$\\
  \hline
  \hline    
Marching Cubes~\cite{lorensen1987marching} & 10.653 & 1.196 & 0.686 & 0.585 & 0.852 & 0.932 & 0.900 & 0.951 & 0.977  & \textbf{0.001} & \textbf{0.005} & \textbf{0.03} \\
Dual Contouring~\cite{ju2002dual}          & 5.992 & 0.814 & 0.648 & 0.758 & 0.911 & 0.937 & 0.923 & 0.961 & 0.979 & 0.09 & 0.7 & 5.6 \\
VoronoiNet~\cite{williams2020voronoinet}   & 3.252 & - & - & 0.543 & - & - & 0.886 & - & - & 11.6 & - & - \\
Shape As Points~\cite{peng2021shape}       & 6.543 & 1.906 & 0.669 & 0.589 & 0.857 & 0.934 & 0.894 & 0.949 & 0.978 & 51.2 &106.3 &191.3  \\
VoroMesh                                   & \textbf{0.791} & \textbf{0.645} & \textbf{0.634} & \textbf{0.920} & \textbf{0.938} & \textbf{0.939} & \textbf{0.958} & \textbf{0.975} & \textbf{0.982} & 2.0 & 4.1 & 36.3  \\
\hline
\end{tabular}}
    \caption{Quantitative comparisons of Chamfer distance (CD), F1 score, and normal consistency (NC) of \textit{VoroMesh} for an optimization-based 3D reconstruction task on the Thingi32 dataset for three different grid resolutions.}
    \label{tab:direct}
  \end{center}
\end{table*}

We implemented our VoroMesh approach in Python with PyTorch~\cite{paszke19neurips}, SciPy~\cite{virtanen20methods} and CGAL~\cite{cgal}. All timings were computed on a Dell Precision desktop machine with 54 cores, a NVidia A6000 GPU, and 512GB of RAM. Our code can be found online at \href{https://nissmar.github.io/voromesh.github.io/}{our project page}, along with most of our results.

\subsection{Optimization-based 3D Reconstruction}
We first compare the ability of different representations --- including our \textit{VoroMesh} --- to capture ground truth surfaces by fitting them with direct optimization. For all methods, we assume that we have access to the ground truth meshes from Thingi32 dataset, which is a subset of Thingi10K~\cite{zhou2016thingi10k}, from which we can derive signed distance values and sample dense point clouds. We exclude two non-watertight models (96481 and 58168) to obtain a final dataset composed of 30 valid shapes. We evaluate our method using grid resolutions of sizes $g_s^3$ with $g_s \!\in\! [32, 64, 128]$. We sample the ground truth surface with $150 \times g_s^2$ points, and use grid vertices adjacent to sampled points as initial Voronoi generators. To boost the optimization, we randomly select $20\%$ of the surface points $X$ that we feed in the \textit{VoroLoss} and use 400 optimization steps of the Adam optimizer~\cite{kingma15iclr} with a learning rate of $0.005$, further divided by two at steps 80, 120, 200 and 250. We timed MC~\cite{lorensen1987marching} and DC~\cite{ju2002dual} (excluding the SDF extraction step), and measured the optimization time (which is systematically the longest step) for the rest of the methods (see supplementary material for timings). \smallskip

\mypar{Baselines}
As no public implementation is available, we re-implemented the soft-Voronoi loss from VoronoiNet~\cite{williams2020voronoinet}. We simplified their pipeline by assigning the correct occupancy to each generator prior to optimization in order to focus only on the ability to fit the surface and ignore the two additional losses (\textit{bounds loss} and \textit{signed-distance loss}) designed to relocate the generators to their assigned regions (inside or outside). In addition, we do not utilize centroidal Voronoi loss for two reasons: (1) it requires computing the full Voronoi diagram at each iteration (which we avoid to keep computations low); (2) and because we start with a uniform grid, so the generators are already well distributed. We select only the generators close to the input mesh to be able to use the same number of samples as for our method.\smallskip

\begin{figure}[t]
\captionsetup[subfigure]{aboveskip=-1pt,belowskip=-1pt}
 \centering
\colimage{.15}{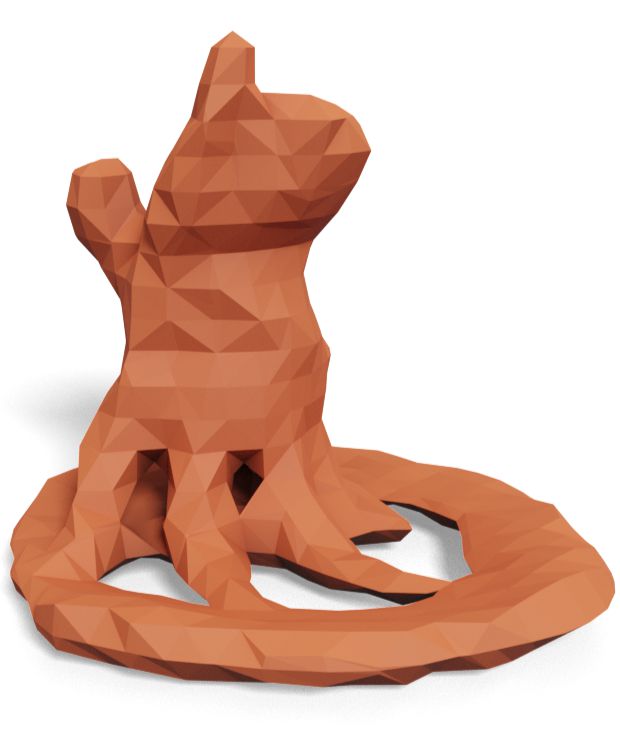}{\caption{MC \cite{lorensen1987marching}}}
\colimage{.15}{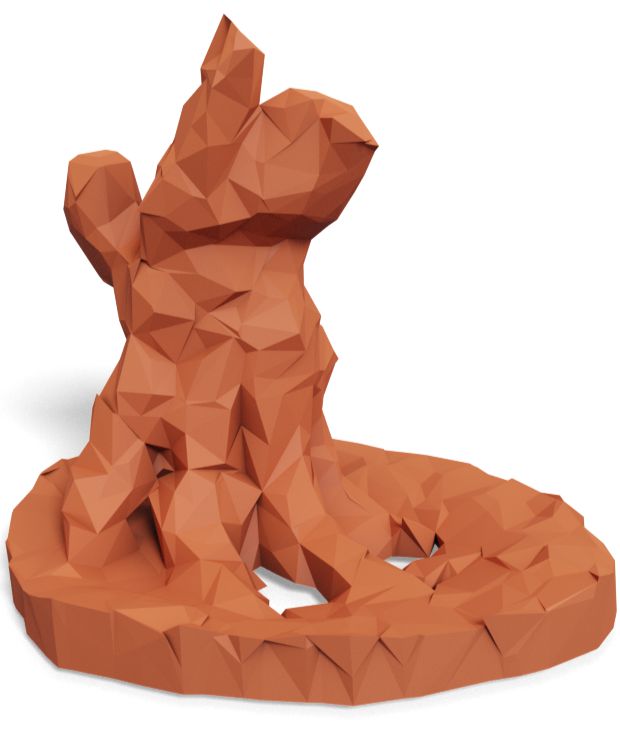}{\caption{DC \cite{ju2002dual}}}
\colimage{.15}{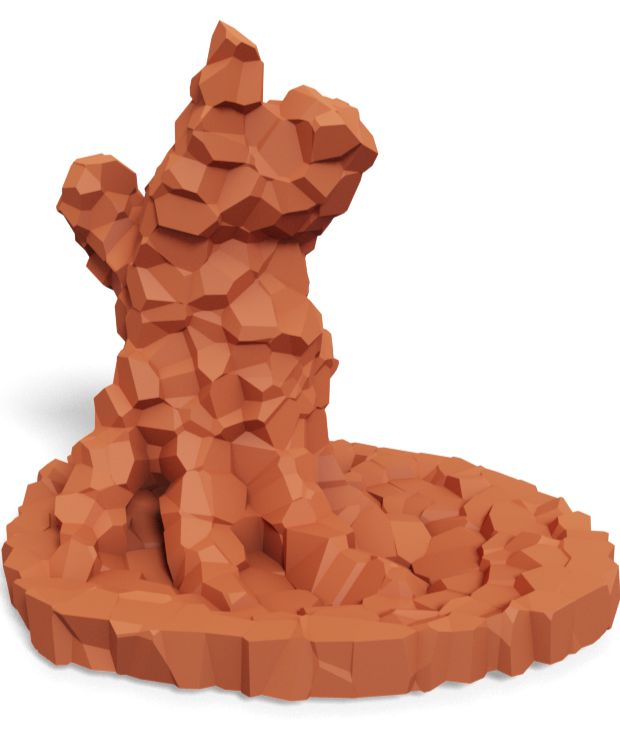}{\caption{VoronoiNet \cite{williams2020voronoinet}}}
 
\colimage{.15}{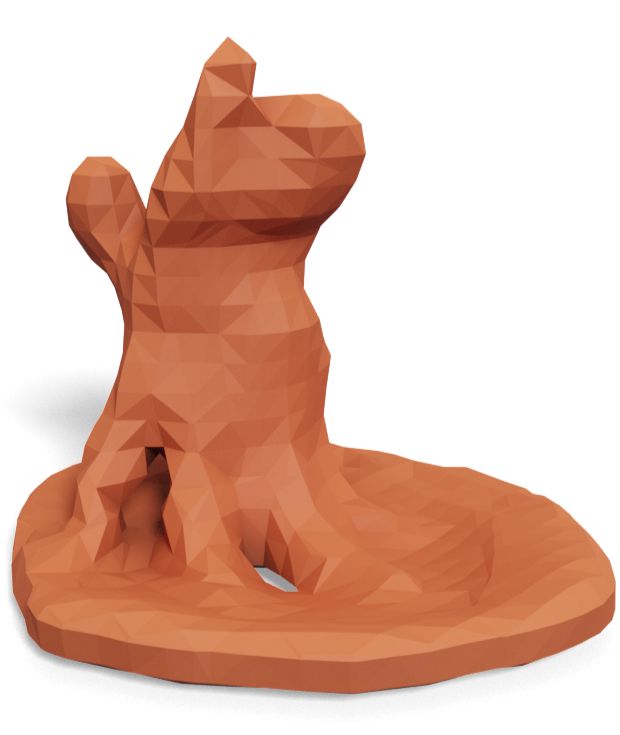}{\caption{SAP \cite{peng2021shape}}}
\colimage{.15}{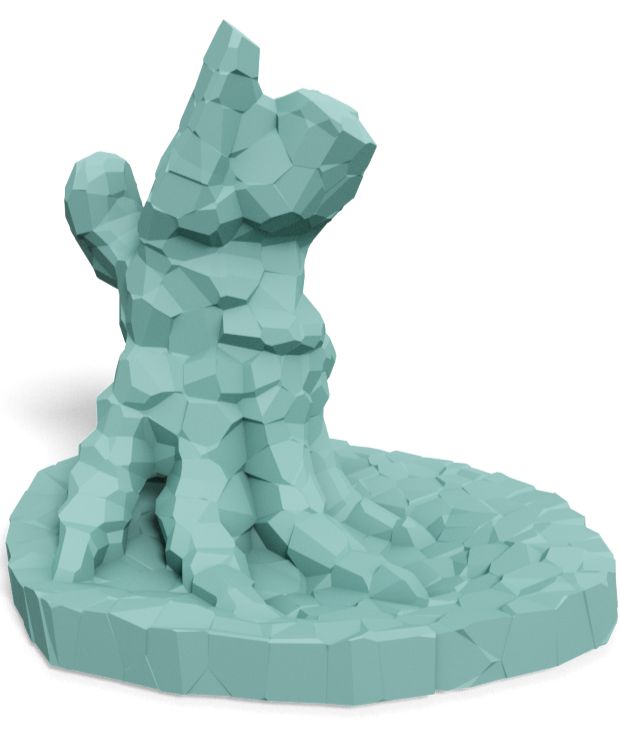}{\caption{Ours}}
\colimage{.15}{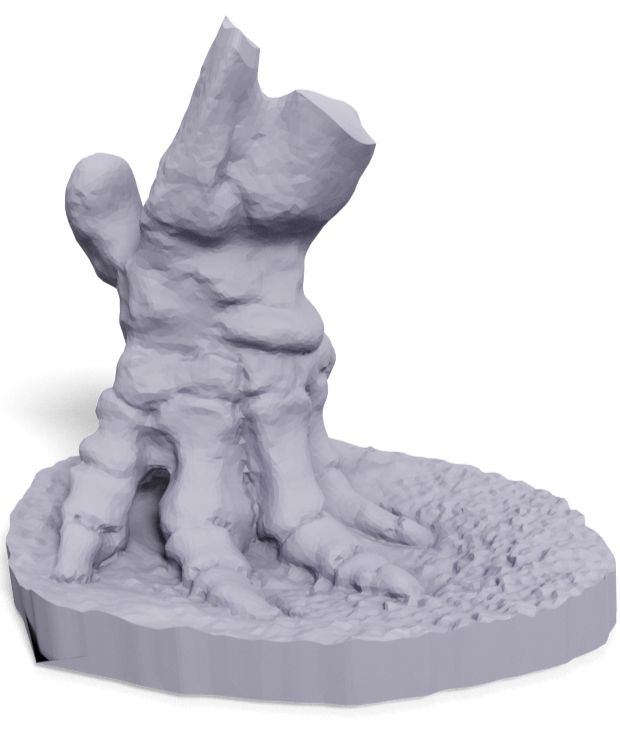}{\caption{Ground Truth}}

  \caption{Visual comparison of optimization-based reconstructions. All methods utilize a grid of size $32^3$.}
  \label{fig:comp32}
\end{figure}

We also compare to Shape-As-Points (SAP) \cite{peng2021shape} which, similarly to \textit{VoroMesh}, relies on the optimization of point positions, using regular grids in the process. SAP performs an iterative refinement scheme with more iterations at each step, giving it a slim advantage over VoronoiNet~\cite{williams2020voronoinet} and our method, but increasing its computation time. We compare to two axiomatic methods as well, namely Marching Cubes (MC)~\cite{lorensen1987marching} and Dual Contouring (DC)~\cite{ju2002dual}. These two methods require a sampling of the SDF as input instead of a point cloud. We evaluate their representation power given the same resolution of the input SDF grid that was used to initialize near-surface generators for \textit{VoroMesh}. We also provide additional comparisons with DMTet~\cite{shen21neurips} in the supplementary material, for completeness.
\\[-2mm]

\mypar{Results} 
Both MC~\cite{lorensen1987marching} and DC~\cite{ju2002dual} suffer from discretization artifacts, missing parts as shown in \fig{comp32} (see supplemental material for examples on more grid sizes), while MC does not reproduce sharp features well.

Originally designed for 2D images, VoronoiNet \cite{williams2020voronoinet} provides a good fit of the overall volume, but a poor fit of the surface normals, which leads to poor visual appearance and NC distance, see \fig{comp32}. Since all the generators are involved in the loss, it does not scale well with increasing grid resolution either: optimization of a grid larger than $32^3$ does not fit into a 50GB GPU. Consequently, our approach is approximately ten times faster, see \tab{direct}.

SAP~\cite{peng2021shape} tends to generate shapes with fewer details and no sharp edges since its reconstruction is based on Marching Cubes. The output mesh, however, is guaranteed to be watertight. SAP requires an optimization of the whole volume, and is thus substantially slower than VoroMesh. 

Our method performs the best across all considered metrics and resolutions given a similar amount of output information (our choice of initial generators around the surface guarantees that we use approximately the same number of cells as MC~\cite{lorensen1987marching} and DC~\cite{ju2002dual}). This confirms that our generators can effectively represent surfaces in 3D and that our \textit{VoroLoss} allows for an efficient optimization of their positions to fit target surfaces. This optimized placement of generators (see \fig{details}), including the ability to move them beyond their initial cells, means that \textit{VoroMesh} can capture fine details, yielding faithful reconstructions even for low numbers of generators.

\subsection{Robustness to Noise}
\begin{figure}[h]
\captionsetup[subfigure]{aboveskip=-1pt,belowskip=-1pt}
  \centering
  \begin{subfigure}{.11\textwidth}
    \centering
    \includegraphics[width=\linewidth]{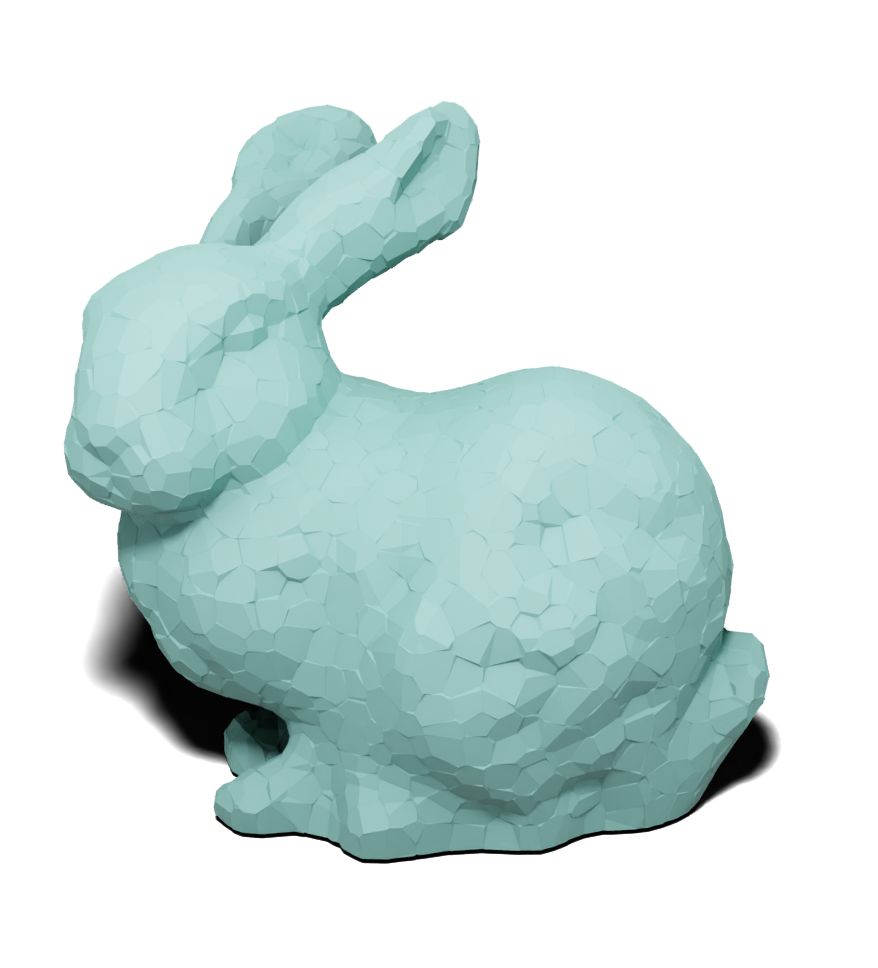}  
    \caption{$\delta = 0\% $}
  \end{subfigure}
  \begin{subfigure}{.11\textwidth}
    \centering    
    \includegraphics[width=\linewidth]{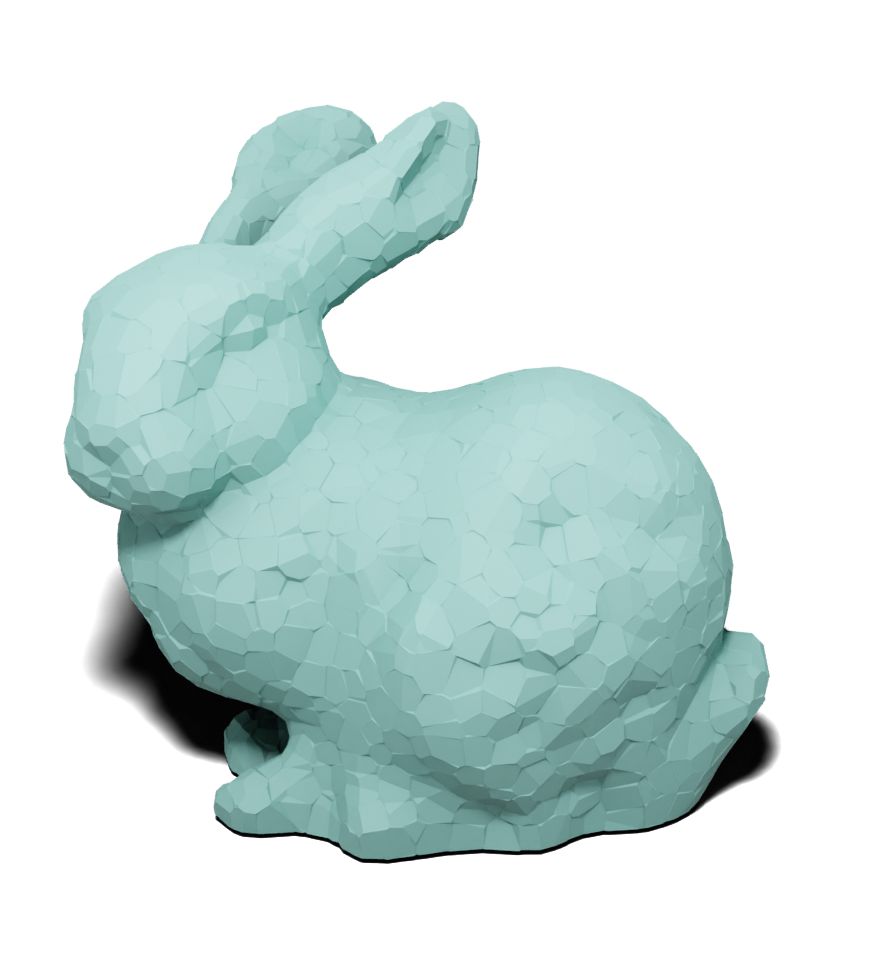}  
    \caption{$\delta = 3.2\% $}
  \end{subfigure}
  \begin{subfigure}{.11\textwidth}
    \centering
    \includegraphics[width=\linewidth]{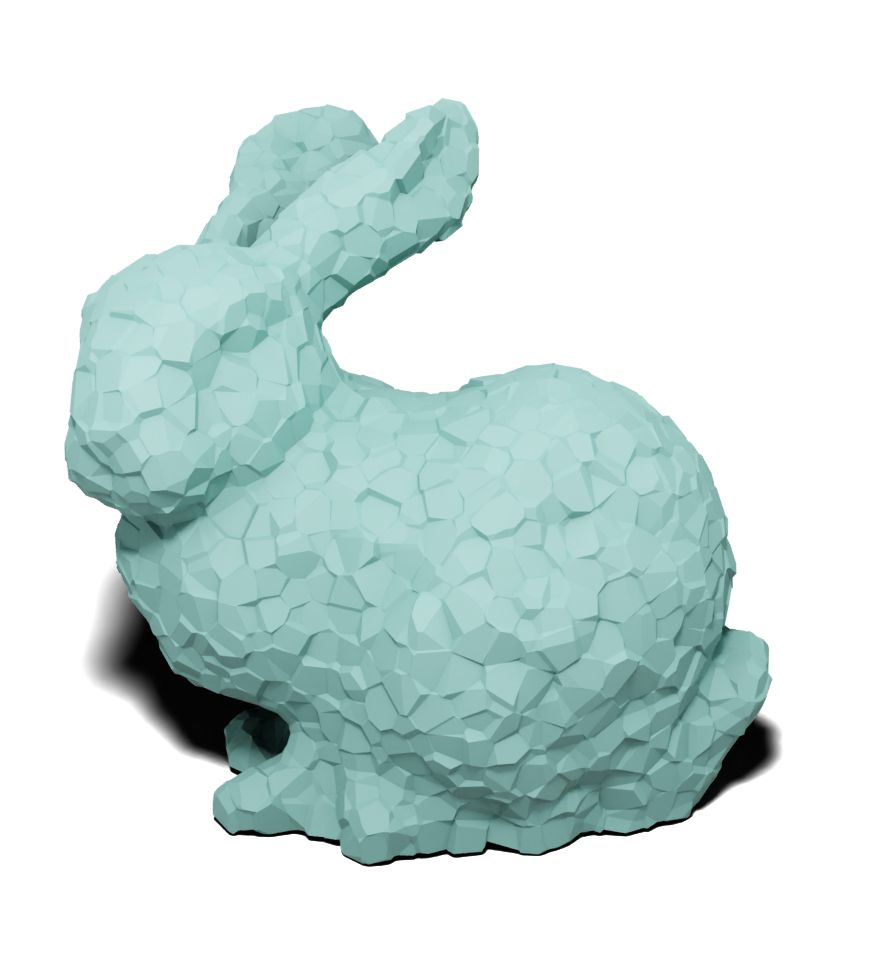}  
    \caption{$\delta = 32\% $}
  \end{subfigure}
  \begin{subfigure}{.11\textwidth}
    \centering
    \includegraphics[width=\linewidth]{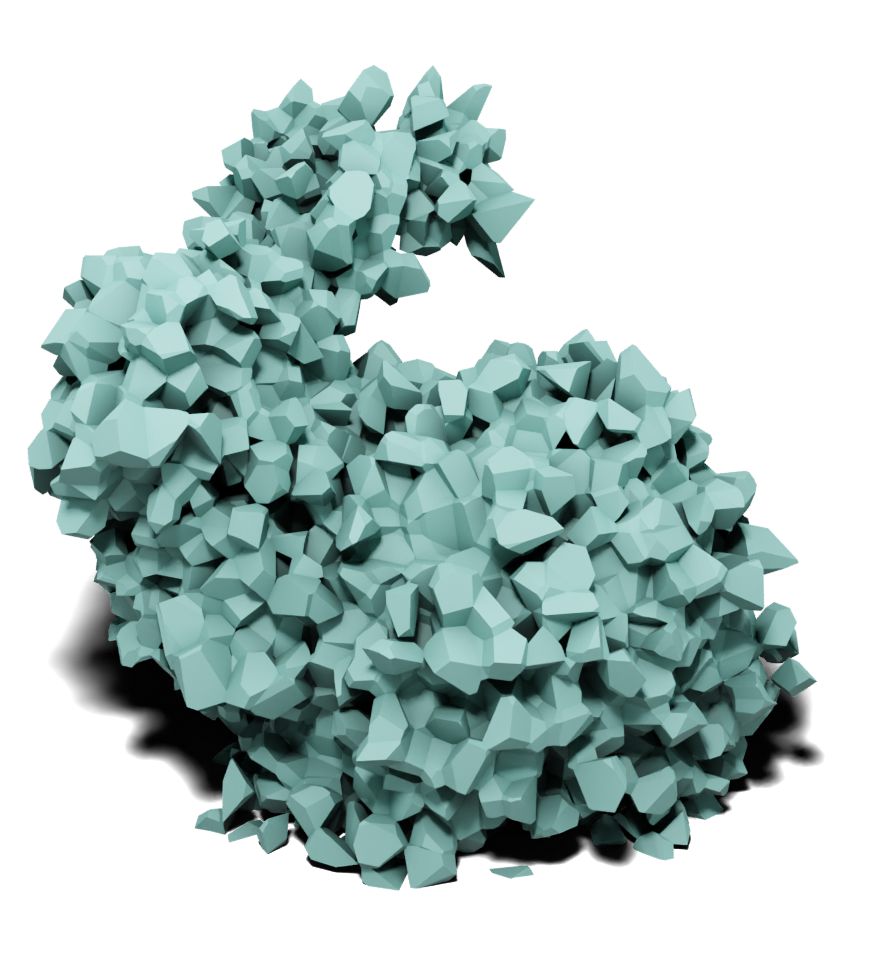}  
    \caption{$\delta = 320\%$}
  \end{subfigure}
  \caption{Robustness to noise. We evaluate the impact of noise on the reconstructed \textit{VoroMesh}, where $\delta$ is the uniform noise magnitude  given as percentage of voxel size.}
  \label{fig:noise}
\end{figure}

In theory, a small displacement of the generators or of their occupancy could induce a large deformation of the \textit{VoroMesh}: for instance, swapping the position of two generators while preserving their occupancy creates a 180-degree flip of the associated face. However, in practice, our representation is robust to small perturbations due to the locality of the Voronoi diagram construction. To demonstrate this property, we applied uniform perturbations to the generators with different magnitudes to one of our reconstructions, see \fig{noise}. This suggests that VoroMesh is suitable for learning-based applications.

\subsection{Learning-based 3D Reconstruction}
We now verify the suitability of \textit{VoroMesh} and \textit{VoroLoss} as target representation and training loss function in an inference-based 3D shape reconstruction from input SDF grids. All methods were trained on the ABC dataset~\cite{koch2019abc}, composed of surface triangle meshes of CAD models. We use the train/validation split from~\cite{chen22tog,chen21tog}, with 4,280 models in the training set and 1,071 models in the testing set. We remove approximately 10\% of non-watertight models in both sets. We additionally verify the generalization capabilities of our method by reconstructing shapes from the Thingi32 dataset \textit{without any fine-tuning}. For this experiment, we consider two grid resolutions, $32^3$ and $64^3$, for which we trained three models using batches of $32$, $16$, and $24$ shapes (one model for each resolution and one using both) and $10^5$ ground truth sample points per shape with the AdamW optimizer~\cite{loshchilov19iclr} and step-wise scheduling of momentum, learning rate, and regularization weight $\lambda$ (see \href{https://nissmar.github.io/voromesh.github.io/}{project page} for code). We first train the SCNN and MLP predicting generators of \textit{VoroMesh} for $200$ epochs, then reuse the convolutional features and train a separate MLP predicting occupancy for each generator for $75$ epochs. \smallskip

\begin{table}[t]
  \begin{center}
    \resizebox{1.\columnwidth}{!}{\begin{tabular}{|l|c|c|c|c|c|c|}
  \hline
  \textbf{Method}            & Grid Size & CD ($\times 10^{-5}$) & F1 & NC & Watertight   \\
  \hline
  \hline
  NDC~\cite{chen22tog}  & $32^3$    & 66.08          & 0.787          & 0.941      & 31\%    \\
  NMC~\cite{chen21tog}  & $32^3$    & 60.70          & 0.833          & \textbf{0.954} & 25\%\\
  VoroMesh32            & $32^3$    & 13.15          & 0.819          & 0.933    & \textbf{100}\%      \\
  VoroMesh32+64         & $32^3$    & \textbf{2.23}  & \textbf{0.835} & 0.941  & \textbf{100}\%           \\
  \hline
  \hline
  NDC~\cite{chen22tog}  & $64^3$    & 2.209          & 0.882          & 0.975    & 20\%      \\
  NMC~\cite{chen21tog}  & $64^3$    & 2.144        & \textbf{0.891} & \textbf{0.980}  & 18\%  \\
  VoroMesh64            & $64^3$    & 1.317          & 0.882          & 0.962   & \textbf{100}\%        \\
  VoroMesh32+64         & $64^3$    & \textbf{1.199} & 0.886          & 0.966    & \textbf{100}\%       \\
  \hline
\end{tabular}}
    \caption{Comparison of learning-based 3D reconstruction methods on the ABC dataset.}
    \label{tab:learning_ABC}
  \end{center}
\end{table}

\mypar{Baselines}
We compare our method against state-of-the-art NMC~\cite{chen21tog} and NDC~\cite{chen22tog} and use their pre-trained models, normalization, and evaluation code to recompute the results on the filtered datasets. Note that NDC proposes an unsigned, augmented version of their algorithm (UNDC); however, the underlying mesh representation does not correspond to the boundary of a volume, as it can collapse thin regions and predict zero thickness surfaces --- see \fig{details} --- so we do not compare to this variant. The Voronoi diagram could also be potentially leveraged to represent a triangle soup: the occupancy information would then lie on the (dual) Delaunay edges rather than Voronoi cells. Since we focus on generating topologically correct surfaces in this work, we leave this possibility for future work. \smallskip

\begin{figure}[t]
 \centering
\captionsetup[subfigure]{aboveskip=4pt,belowskip=0pt}
\colimage{.115}{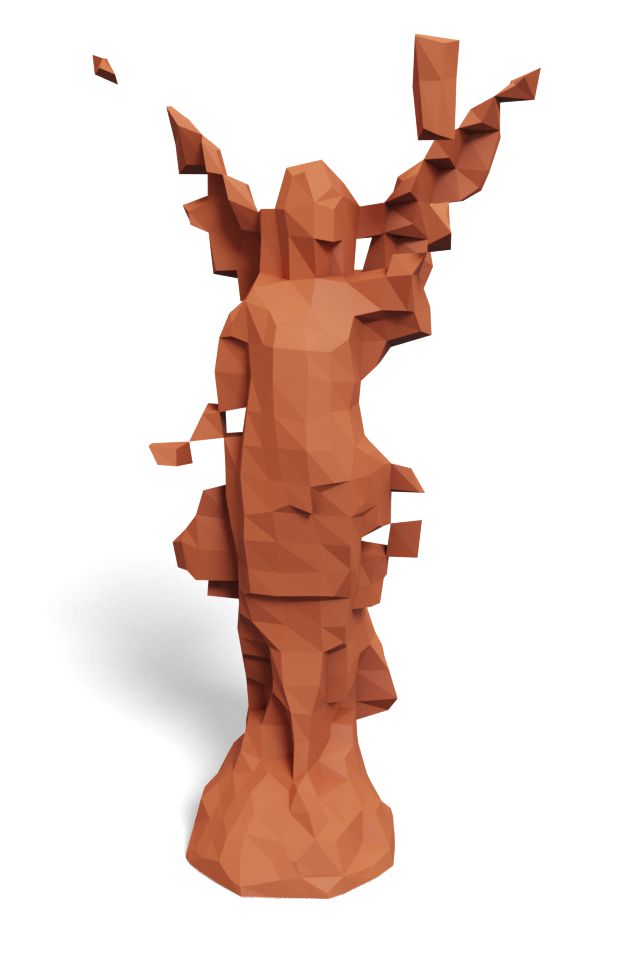}{}
\colimage{.115}{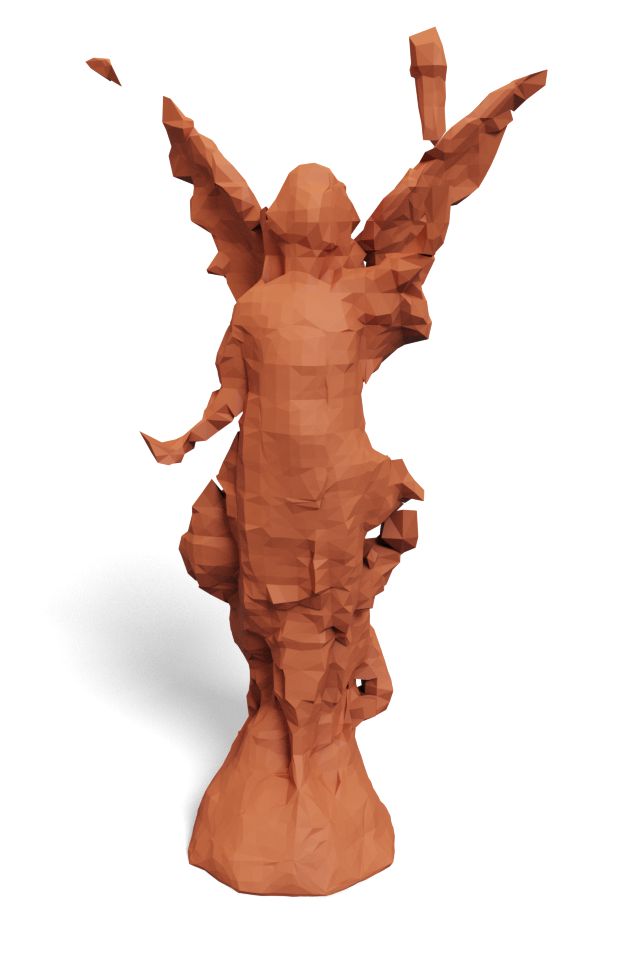}{}
\colimage{.115}{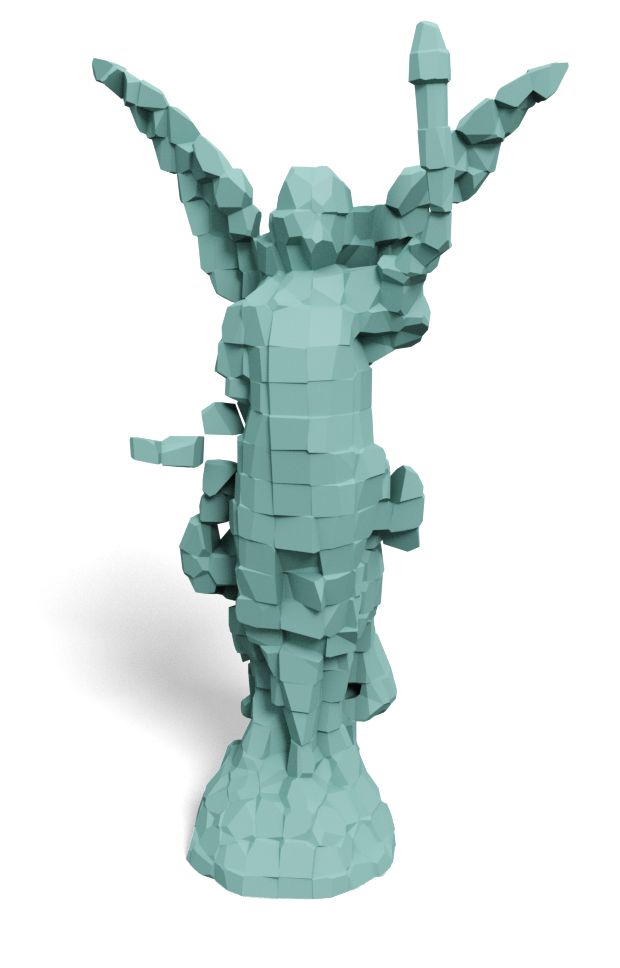}{}
\colimage{.115}{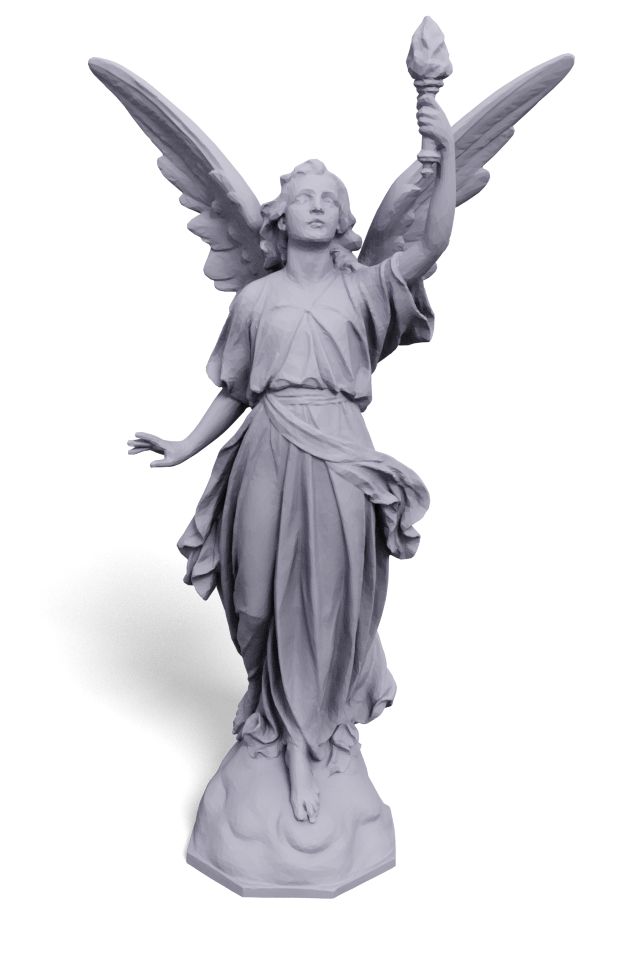}{}\\[-5.5mm]
\colimage{.115}{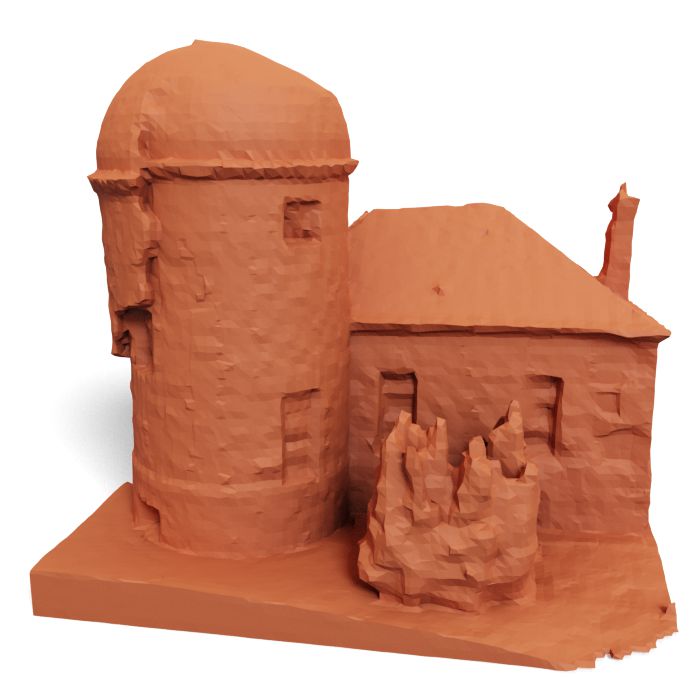}{}
\colimage{.115}{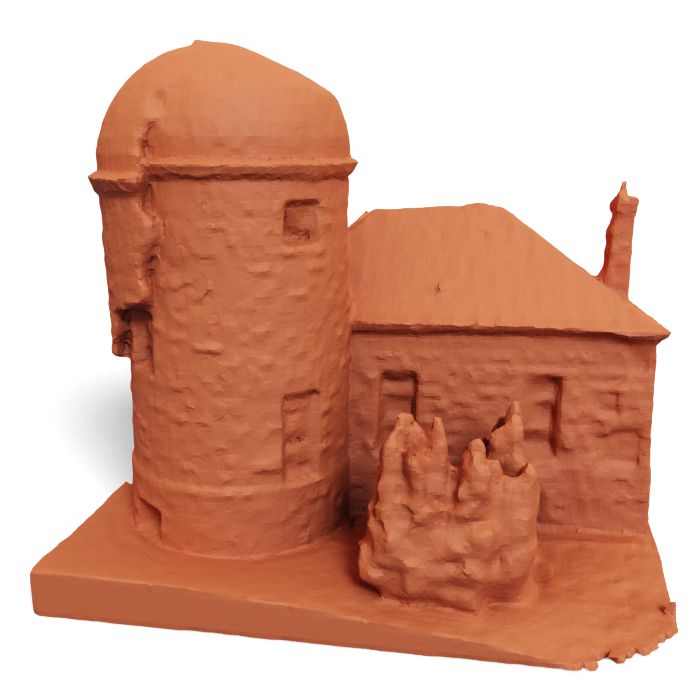}{}
\colimage{.115}{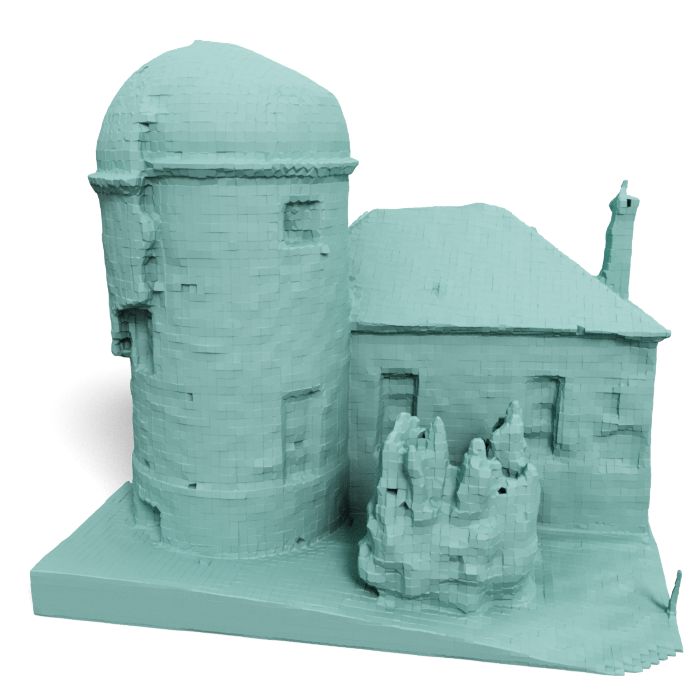}{}
\colimage{.115}{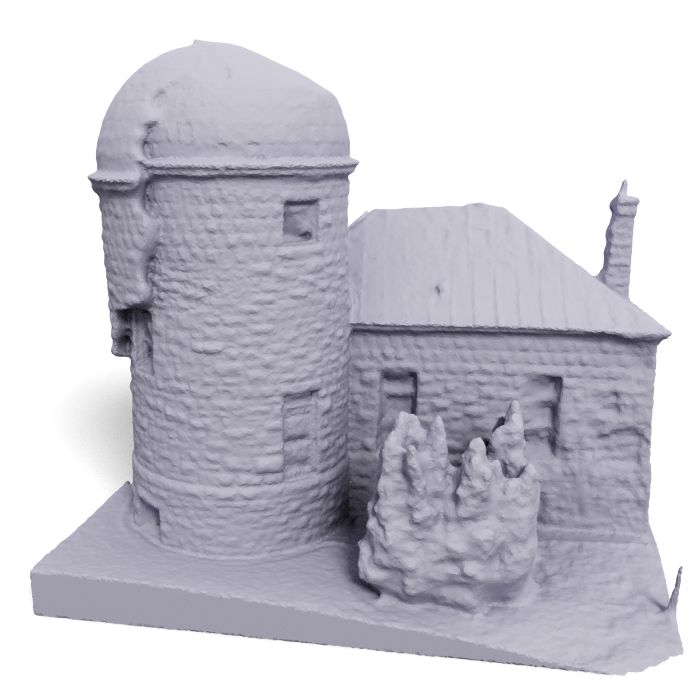}{}\\[-3mm]
\colimage{.115}{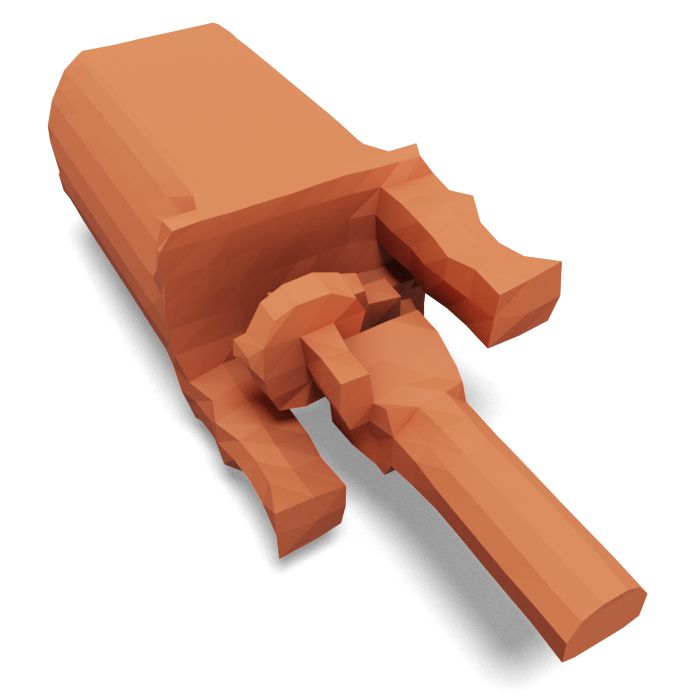}{}
\colimage{.115}{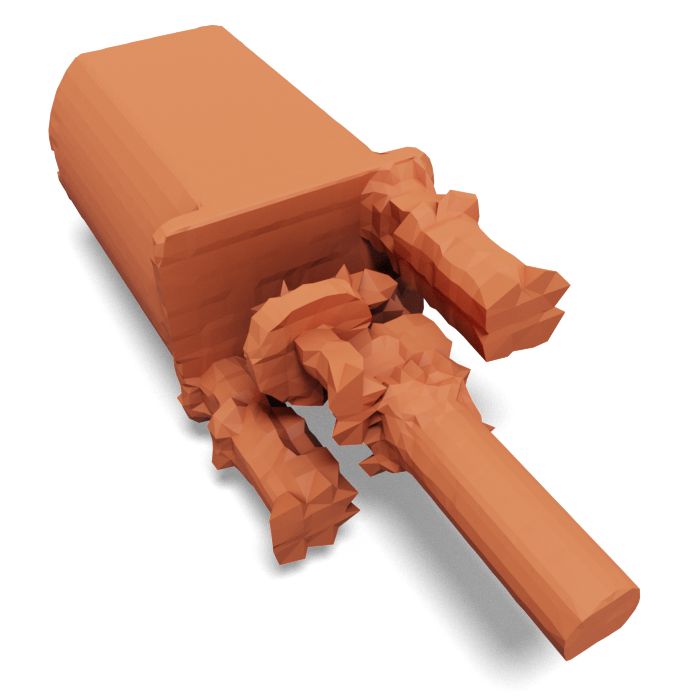}{}
\colimage{.115}{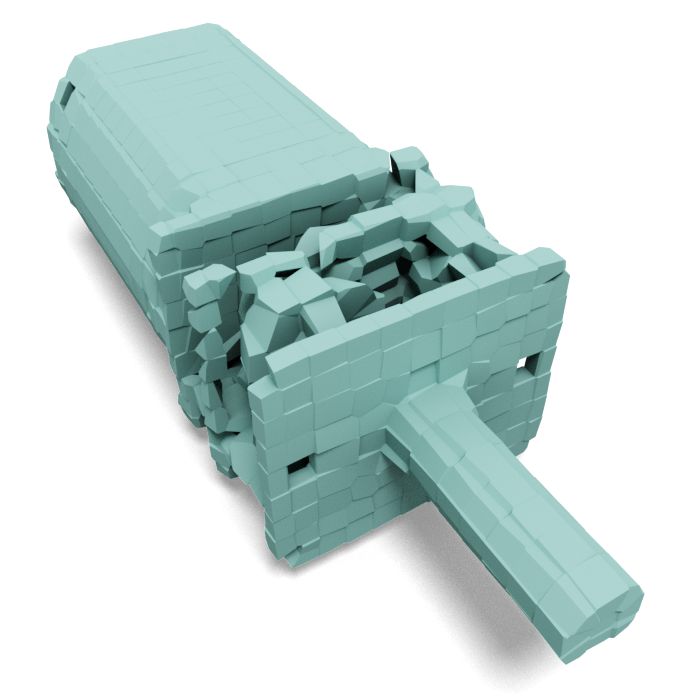}{}
\colimage{.115}{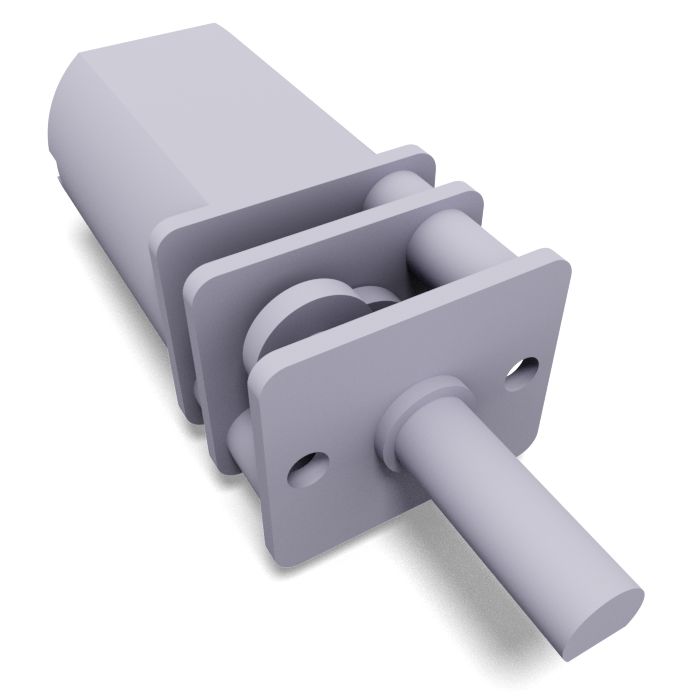}{}\\[-4mm]
\colimage{.115}{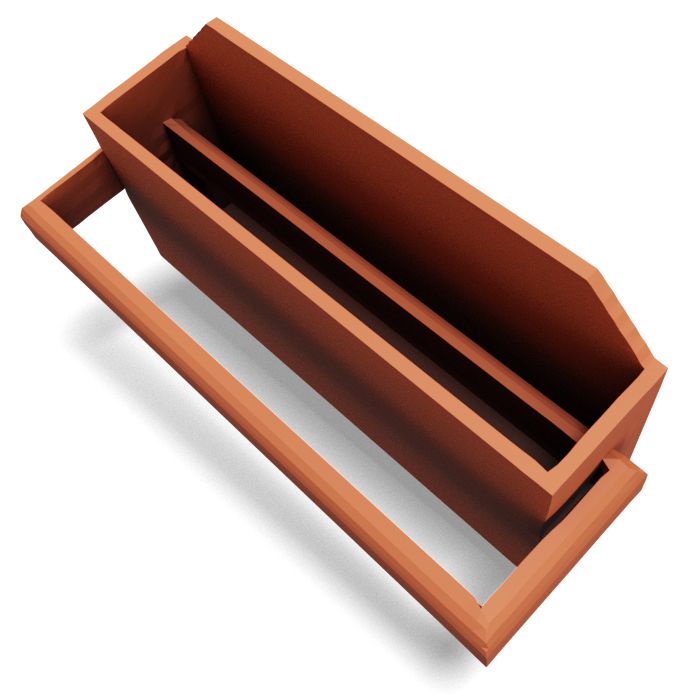}{\caption{NDC~\cite{chen22tog}}}
\colimage{.115}{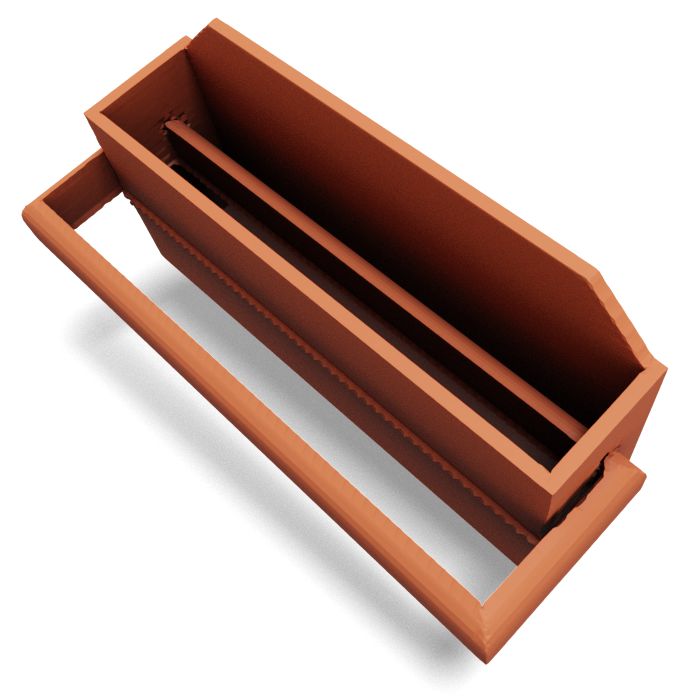}{\caption{NMC~\cite{chen21tog}}}
\colimage{.115}{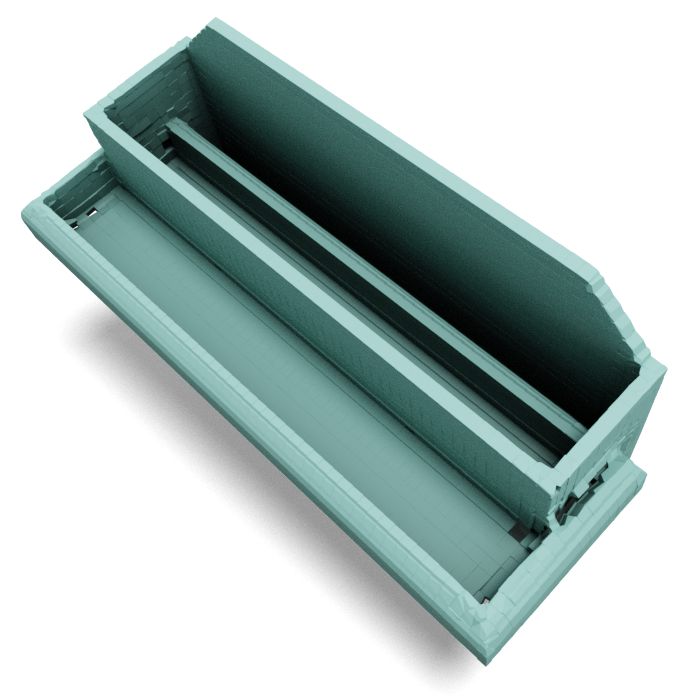}{\caption{VoroMesh}}
\colimage{.115}{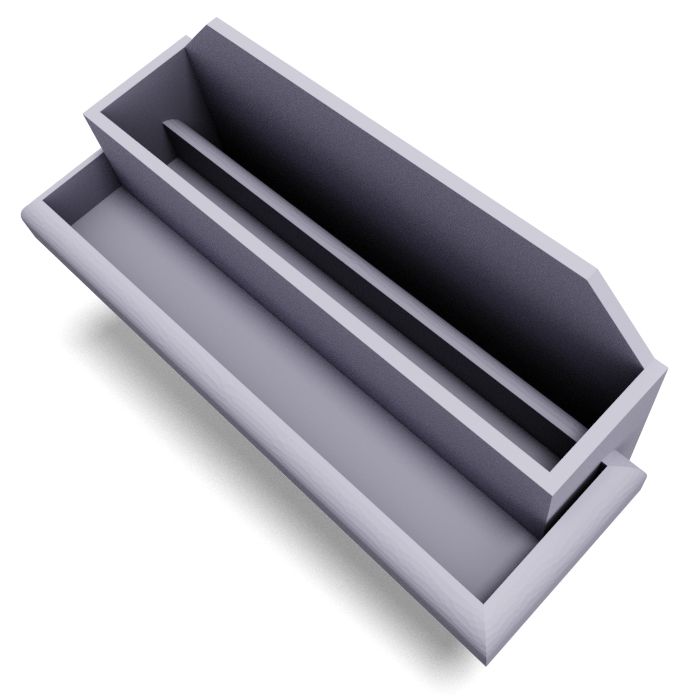}{\caption{GT}}
  \caption{Visual comparison of learning-based reconstruction. Input grids of sizes $32^3$ and $128^3$ for Thingi32 (top rows), $32^3$ and $64^3$ for ABC (bottom rows). Our VoroMesh reconstruction, relying on a model trained on $32^3$ SDF inputs, is able to reconstruct thin details.}
    \label{fig:complearning}
\end{figure}

\mypar{Results}
\textit{VoroMesh} outperforms both baselines in terms of Chamfer distance (CD), while being comparable in terms of normal consistency (NC) and F1 score, see \tab{learning_ABC}. We believe that these results are due to NDC and NMC reconstructing overall smoother surfaces, but failing to capture parts that \textit{VoroMesh} can resolve with proper placement of the generators. NMC, which relies on a custom tessellation, uses more vertices, and thus has a slight advantage in terms of geometric fidelity, as explained by its authors~\cite{chen22tog}. Note that both NMC and NDC fail to produce \textit{any mesh at all} for certain thin shapes from the test set. We also check for watertightness using CGAL~\cite{cgal} by confirming the absence of self-intersection and verifying that each edge is incident to exactly two faces: as expected, all VoroMeshes are watertight while only a fraction of NDC and NMC meshes are.\\[-2mm]

\mypar{Generalization to Thingi32}
To test the ability of our model to transfer knowledge across datasets and verify the stability of our \textit{VoroMesh} representation, we reconstruct the same meshes from the Thingi32 dataset used in our direct optimization experiment, with models pretrained on the ABC dataset. As can be seen in \fig{complearning}, our approach systematically produces valid detailed reconstructions of unseen shapes. 
On a coarse resolution, our approach manages to outperform baselines in terms of CD by a significant margin, meaning \textit{VoroMesh} can capture more details with similar shape parameter budgets. On higher resolutions, NMC slightly pulls ahead, see \tab{learning_thingi}; 
we speculate that it is due to the fact that Thingi32 contains fewer thin structures, so successful reconstructions of such shapes do not demonstrate the ability to capture such details. 
Note that all three methods outperform classical MC across all resolutions and, in some cases, DC (although it uses richer inputs with normals at edge intersection points): with the same SDF inputs, neural methods are able to recover more information.

\begin{table}[t]
  \begin{center}
    \resizebox{1.\columnwidth}{!}{

\begin{tabular}{|l|c|c|c|c|r|}
  \hline
  \textbf{Method}            & Grid Size & CD ($\times 10^{-5}$) & F1 & NC  & Watertight  \\
  \hline
  \hline
  NDC~\cite{chen22tog}  & $32^3$  & 6.396          & 0.734          & 0.918    &   10\%    \\
  NMC~\cite{chen21tog}  & $32^3$  & 5.212          & \textbf{0.796} & \textbf{0.936} & 0\% \\
  VoroMesh32           & $32^3$  & \textbf{2.146} & 0.738          & 0.898       & \textbf{100}\%   \\
  VoroMesh32+64        & $32^3$  & 2.833          & 0.758          & 0.902   & \textbf{100}\%       \\
  \hline
  \hline
  NDC~\cite{chen22tog}  & $64^3$  & 0.863          & 0.906          & 0.960        & 0\%  \\
  NMC~\cite{chen21tog}  & $64^3$  & \textbf{0.779} & \textbf{0.923} & \textbf{0.969} & 0\%\\
  VoroMesh32           & $64^3$  & 0.868          & 0.901          & 0.934    & \textbf{100}\%      \\
  VoroMesh32+64        & $64^3$  & 1.031          & 0.906          & 0.939    & \textbf{100}\%      \\
  \hline
  \hline
  NDC~\cite{chen22tog}  & $128^3$ & 0.651          &  0.937         & 0.980       & 0\%   \\
  NMC~\cite{chen21tog}  & $128^3$ & \textbf{0.642} & \textbf{0.938} & \textbf{0.984} & 0\% \\
  VoroMesh32           & $128^3$ & 0.659          & 0.935          & 0.956     & \textbf{100}\%     \\
  VoroMesh32+64        & $128^3$ & 0.731          & 0.932          & 0.959     & \textbf{100}\%     \\
  \hline
\end{tabular}}
    \caption{Quantitative comparisons with \textit{VoroMesh} for inference-based 3D mesh reconstruction on the Thingi32 dataset, using input SDF grids of different resolutions.}
    \label{tab:learning_thingi}
  \end{center}
\end{table}

\section{Conclusion}
\label{sec:conclusion}

In this paper, we presented VoroMesh, a differentiable Voronoi-based representation of watertight 3D shape surfaces along with its VoroLoss. 
As the resulting watertight surface mesh is defined uniquely (but implicitly) by a Voronoi diagram of optimized generators, we showed that our representation provides significant improvement in geometric fidelity compared to previous works, and captures small or sharp features well. 
We demonstrated how our VoroLoss applies to either direct optimization of generators, or to the training of a neural network for prediction of generators. 
While we assume for simplicity of exposition that our ground truth inputs for generator optimization are triangle meshes, note that any format supporting inside/outside queries would do as well --- even point sets, where, e.g., the generalized fast winding number \cite{barill2018fast} can be used to extract  occupancy information. \smallskip

Our work offers a number of further research directions. First, the VoroMesh could be post-processed to remove the small faces that can cause surface artifacts in the final mesh, in particular for low-resolution VoroMeshes (see supplementary material for additional details). Second, our initialization of generators has the advantage of being simple, but there may be more efficient ways to proceed, even at the risk of having initially too many generators; if some of them do not affect the VoroMesh, they can simply be discarded later. Third, surface-based regularization may provide a mechanism to further improve preservation of sharp features of the input in the VoroMesh. Finally, our representation can potentially be combined with recent advances in generative modeling of point clouds, in order to produce global mesh prediction models that are suitable for generative applications with guarantees of the topological correctness of output meshes.

\paragraph*{Contributions.} \label{sec:Contributions}
Nissim developed the VoroMesh representation and the VoroLoss, performed the optimization-based experiments and proposed a two-stage learning pipeline compatible with this representation. Roman boosted the performances of the VoroLoss, developed the neural network architecture and carried out the learning-based reconstruction and generalization experiments. 

\paragraph*{Acknowledgments.} Nissim Maruani and Pierre Alliez are supported by the French government, through the 3IA C\^ote d'Azur Investments in the Future project managed by the National Research Agency (ANR) with the reference number ANR-19-P3IA-0002. Maks Ovsjanikov and Roman Klokov acknowledge ERC Starting Grant No. 758800 (EXPROTEA) and the ANR AI Chair AIGRETTE. Mathieu Desbrun acknowledges the generous support of Ansys and Adobe Research, as well as a Choose France Inria chair.

{\small
\bibliographystyle{ieee_fullname}
\bibliography{egbib}
}

\clearpage

\appendix

\twocolumn[\begin{center}
      {\Large \bf \emph{Supplemental Material for:}\\ VoroMesh: Learning Watertight Surface Meshes with Voronoi Diagrams\\ICCV '23 \par}
      \vspace*{34pt}
\end{center}]

In \sect{voroloss}, we provide a proof of Theorem 1 from our ICCV paper and discuss how our VoroLoss can be implemented efficiently using a $k$-nearest-neighbors algorithm. We then present a detailed analysis of the chamfer distances obtained with our model trained on the ABC dataset in \sect{cdhist}, before demonstrating additional qualitative comparisons for both optimization-based and learning-based results in \sect{qualall}. \sect{artifacts} discusses how to improve the outputs of VoroMesh, while a comparison of our approach to DMT~\cite{shen21neurips} is presented in \sect{DMTet}. Finally, additional timings of our experiments are provided in \sect{timings}.

\section{Additional comments about VoroLoss}\label{sec:voroloss}

\begin{figure}[!h]
 \centering
 \begin{subfigure}{.23\textwidth}
  \centering
  \includegraphics[width=\linewidth,draft=false]{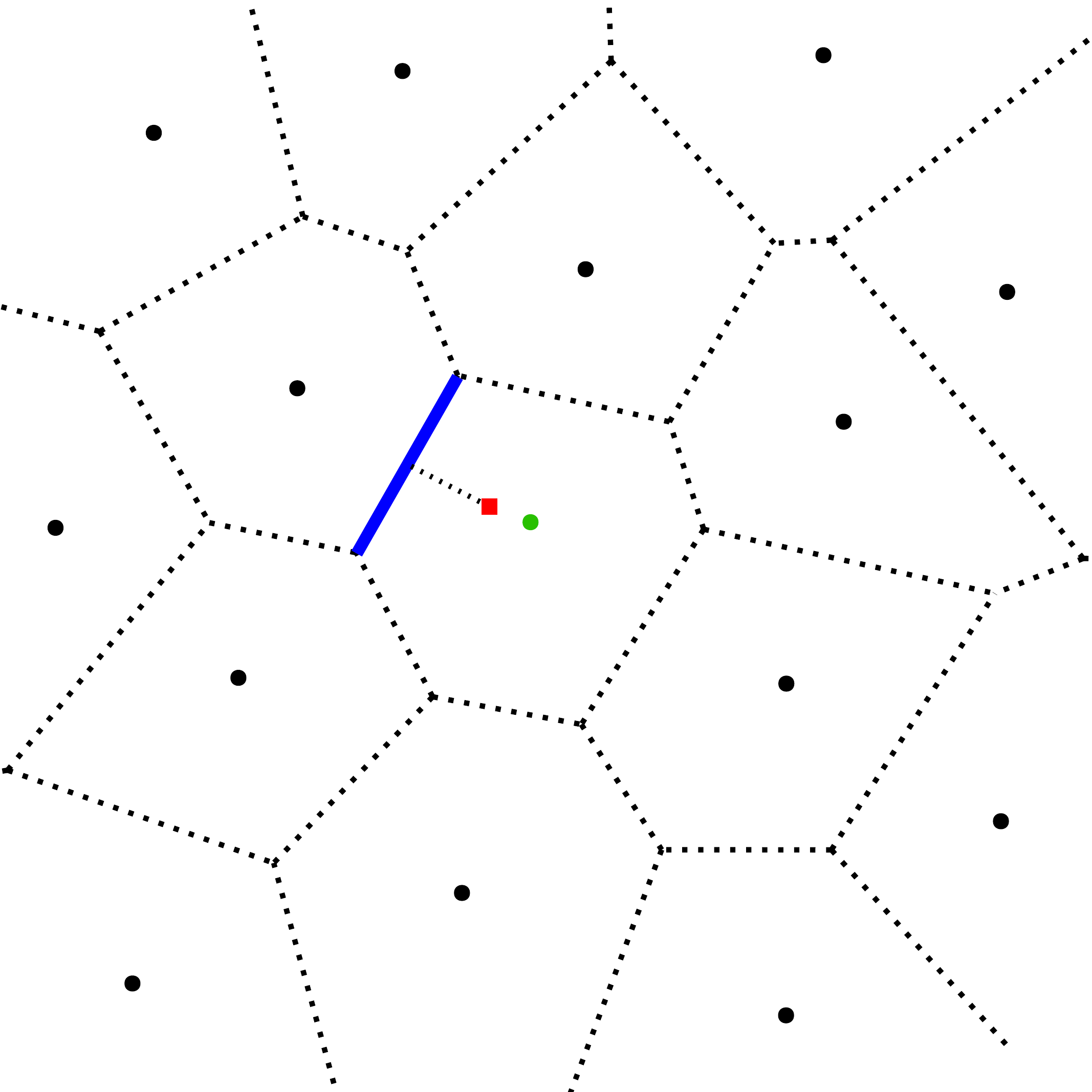}\caption{Closest face}
\end{subfigure}
\begin{subfigure}{.23\textwidth}
  \centering    
   \includegraphics[width=\linewidth,draft=false]{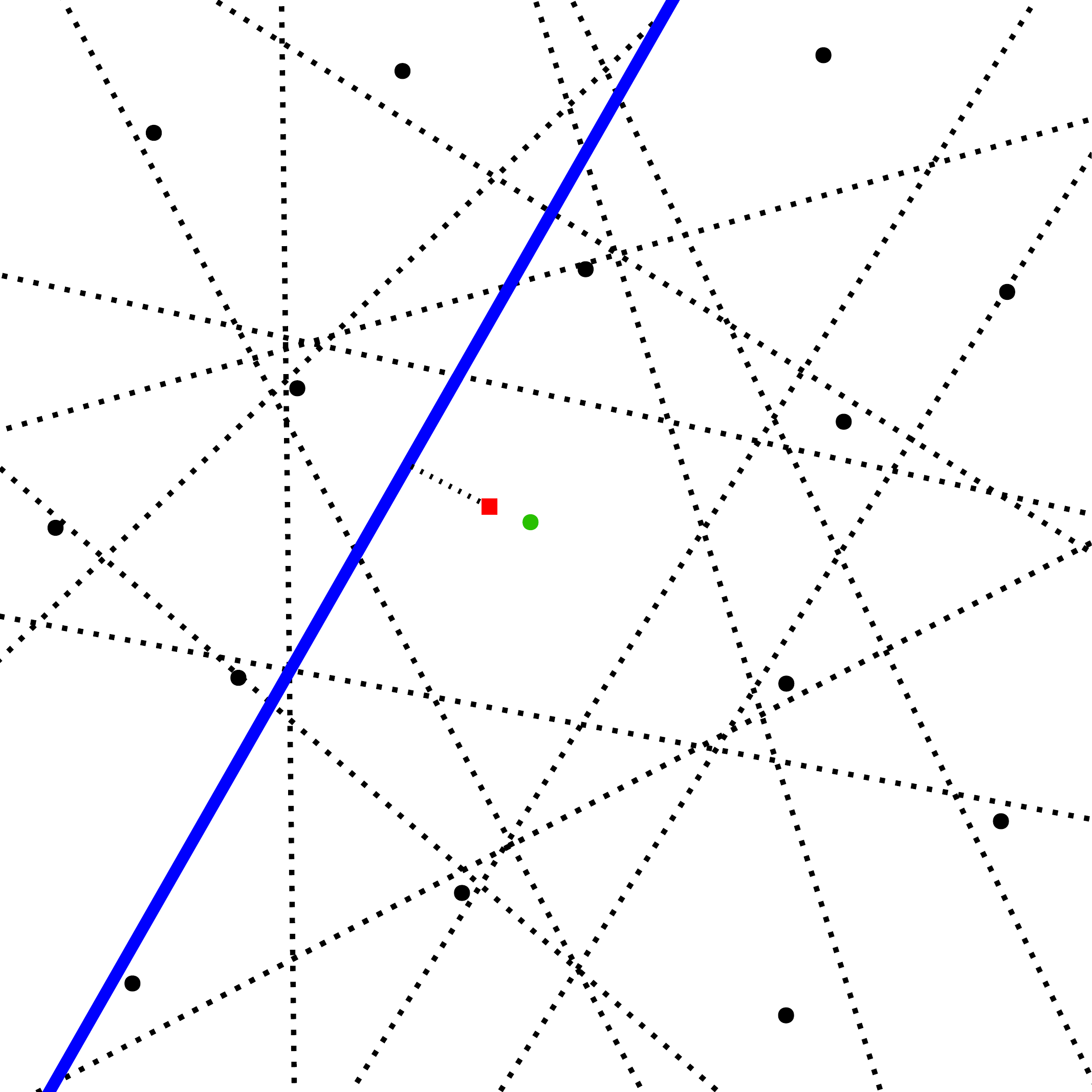}\caption{Closest bisector}
\end{subfigure}
  \caption{Visual proof of our \textit{VoroLoss}: given a point (red) of the target surface, the distance to the closest face in the Voronoi diagram (a) is equal to the distance to the closest bisector between the generator (green, in which the sampled point lies) and all the other ones (b).} 
    \label{fig:proof}
\end{figure}

\mypar{Proof of Theorem 1} To validate our Voroloss, we need to prove that the distance from a point $x \!\in\! \mathbb{R}^3$ to the set of cell faces of the Voronoi diagram of the generators equates the distance from $x$ to the bisector planes between the generator $q_i$ of the Voronoi cell containing $x$ and the other generators $q_{j\neq i}$. 
Note that the Voronoi cell $V_i$ of a generator $q_i$ is the intersection of half-spaces containing $q_i$~\cite{aurenhammer1991voronoi} (see Figure~\ref{fig:proof}), i.e., $V_i \!=\! \cap_{j \neq i} H^i_{i,j}$ where $H^i_{i,j} \!=\! \{ y \!\in\! \mathbb{R}^3 | \|y-q_i\| \!\leq\! \|y-q_j\| \}$ denotes the half-space of points closer to $q_i$ than $q_j$, where the boundary $\partial H^i_{i,j}$ of $H^i_{i,j}$ is the bisector plane $H_{i,j}$ between $q_i$ and $q_j$. Therefore, each Voronoi cell is convex, implying the property we stated since:\vspace*{-1mm}
$$d(x,\partial V_i) = d(x, \cup_{j \neq i} H_{i,j}) = \min_{j\neq i} d(x, H_{i,j}).$$

\begin{figure}[!h]
  \centering
\begin{subfigure}{.23\textwidth}
    \centering
    \includegraphics[width=\linewidth,draft=false]{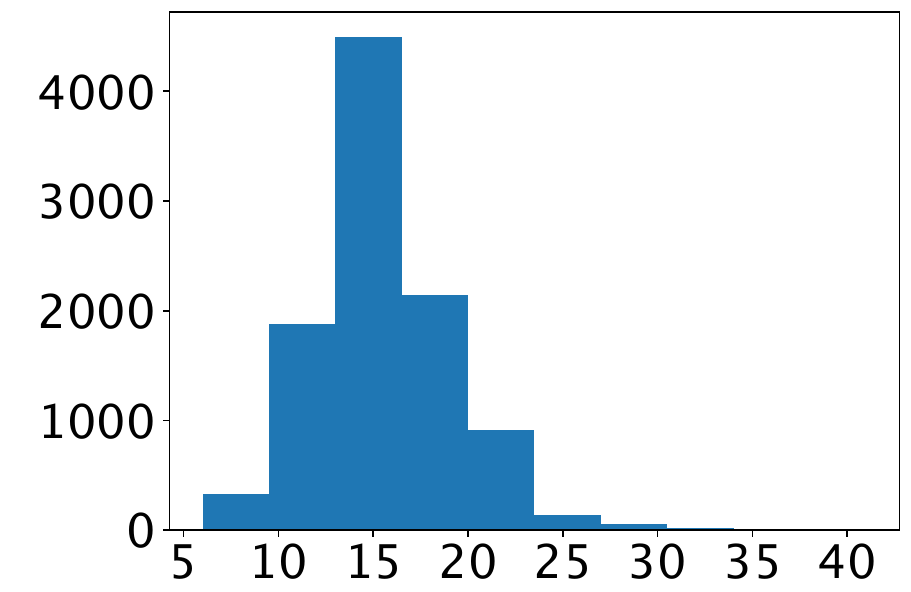}\caption{\#neighbors per Voronoi cell}\label{fig:knnneigh}
  \end{subfigure}
\begin{subfigure}{.23\textwidth}
    \centering   \includegraphics[width=\linewidth,draft=false]{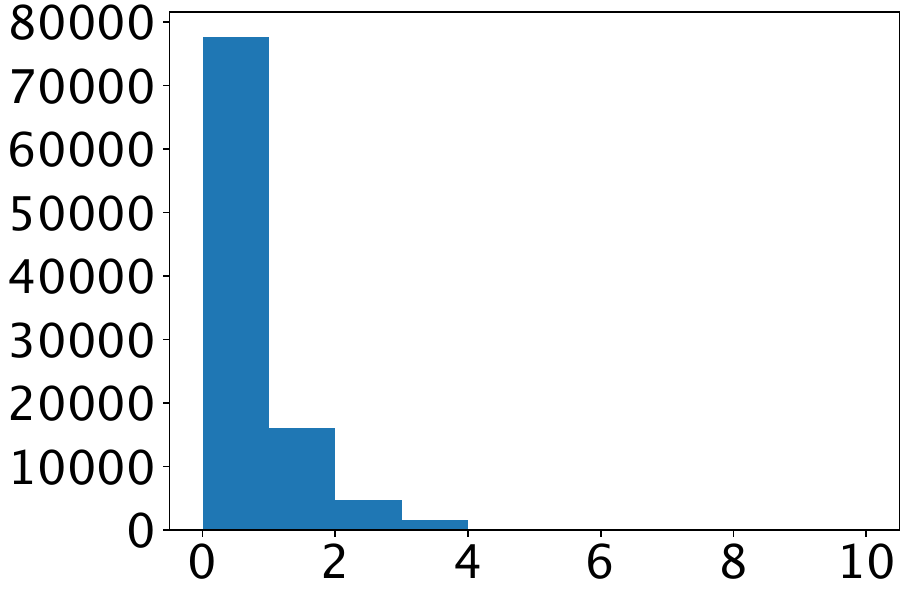}\caption{Index of closest bisector}\label{fig:knn}
  \end{subfigure}
  \vspace*{-2mm}
\caption{Statistics for a random 3D Voronoi diagram.\vspace*{0mm}}\label{fig:stat}
\end{figure}

\mypar{VoroLoss implementation details} Our \textit{VoroLoss} can be efficiently computed using a k-nearest-neighbors algorithm. In this experiment, we consider the Voronoi diagram of $10^5$ points placed randomly in $[-1,1]^3$. Let $i$ be the index of the closest generator for a point $x$ of the dense sampling. The average number of faces for each Voronoi cell (which also corresponds to the degree of each vertex in the dual (Delaunay) triangulation) can be large, see Figure~\ref{fig:knnneigh}. However, when sorting the generators by distance to the sampled point, the $j$-th index of the generator for which $d(x, H_{i,j})$ is minimal is low, see Figure~\ref{fig:knn}. Note that we include an implementation of \textit{VoroLoss} (with the complete architectures of our networks) in the file \verb|voromesh.py| of our supplemental material.

\begin{figure*}[t]
  \centering
  \includegraphics[width=1.9\columnwidth, draft=false]{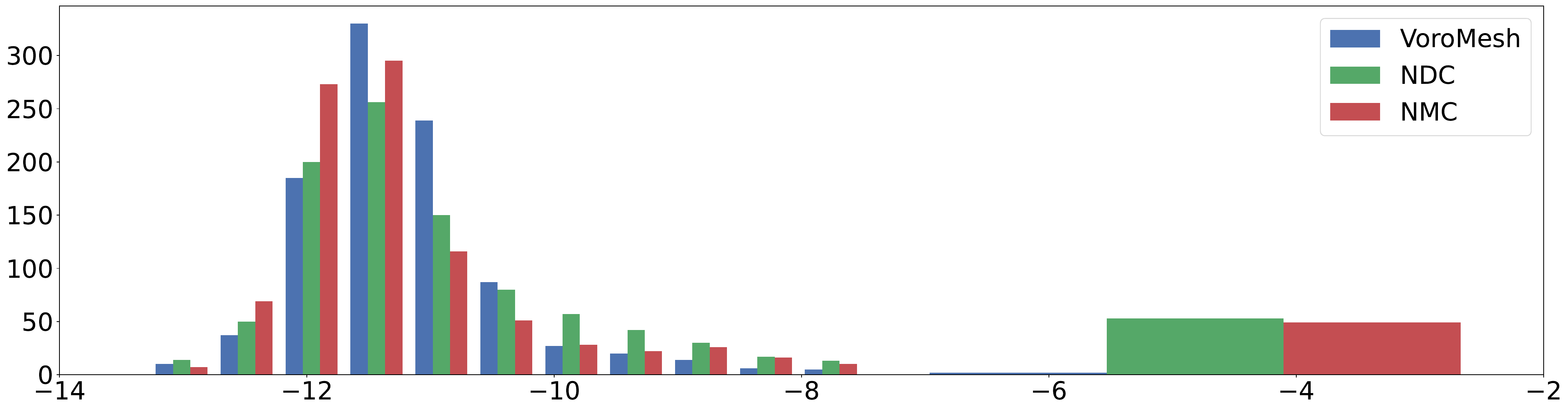} \vspace*{-2mm}
  \caption{Histogram of Chamfer distance log values for VoroMesh,  NMC~\cite{chen21tog}, and NDC~\cite{chen22tog} for SDF inputs of resolution $32^3$.\vspace*{-2mm}}
 \label{fig:cdhist32}
\end{figure*}

\begin{figure*}[t]
  \centering
  \includegraphics[width=1.9\columnwidth, draft=false]{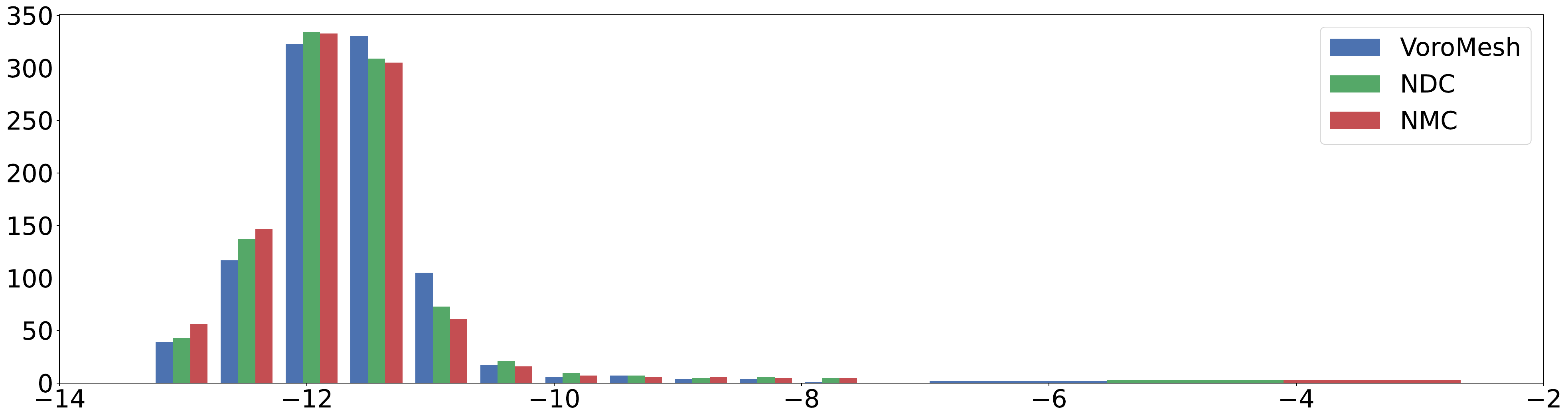} \vspace*{-2mm}
  \caption{Histogram of Chamfer distance log values for VoroMesh,  NMC~\cite{chen21tog}, and NDC~\cite{chen22tog} for SDF inputs of resolution $64^3$.\vspace*{-1mm}}
  \label{fig:cdhist64}
\end{figure*}

\section{Quantitative Analysis}\label{sec:cdhist}
In order to complement our ICCV '23 paper, we provide histograms of chamfer distance (in logarithmic values to exacerbate differences) obtained on the ABC dataset with our \textit{VoroMesh} for input SDF grids of resolutions $32^3$ and $64^3$, and compare it to the state of the art in Figures~\ref{fig:cdhist32}-\ref{fig:cdhist64}. 

Generally, \textit{VoroMesh} produces slightly fewer low-CD reconstructions compared to NMC~\cite{chen21tog} and NDC~\cite{chen22tog}, but it also produces significantly fewer failed reconstructions (i.e., with outlier high-CD values), which allows our method to achieve better aggregate metric values. Our approach is most effective at low resolution, where the previous observation is most obvious; additionally, \textit{VoroMesh} produces significantly more reconstructions in the average quality range.
For higher resolution, \textit{VoroMesh} remains competitive, showing that our loss can be effectively used to produce training signal for detailed high-quality reconstructions. For a grid size $64^3$, VoroMesh still outperforms competitors in aggregate CD values due to its ability to represent finer shapes with adaptive surface discretization and, as a result, its lack of failed reconstructions.

\section{Qualitative Analysis}\label{sec:qualall}

Finally, we present additional renders of the models presented in the articles for all resolutions. We also showcase the different methods on additional models. 

\begin{figure*}[!h]
  \centering
 \colimage{.18}{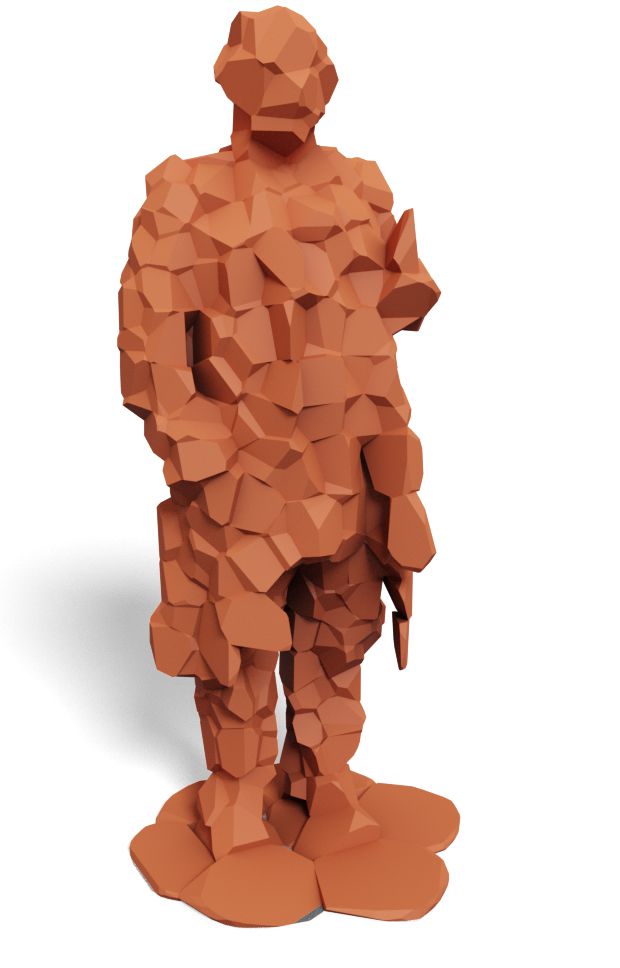}{}
  \colimage{.18}{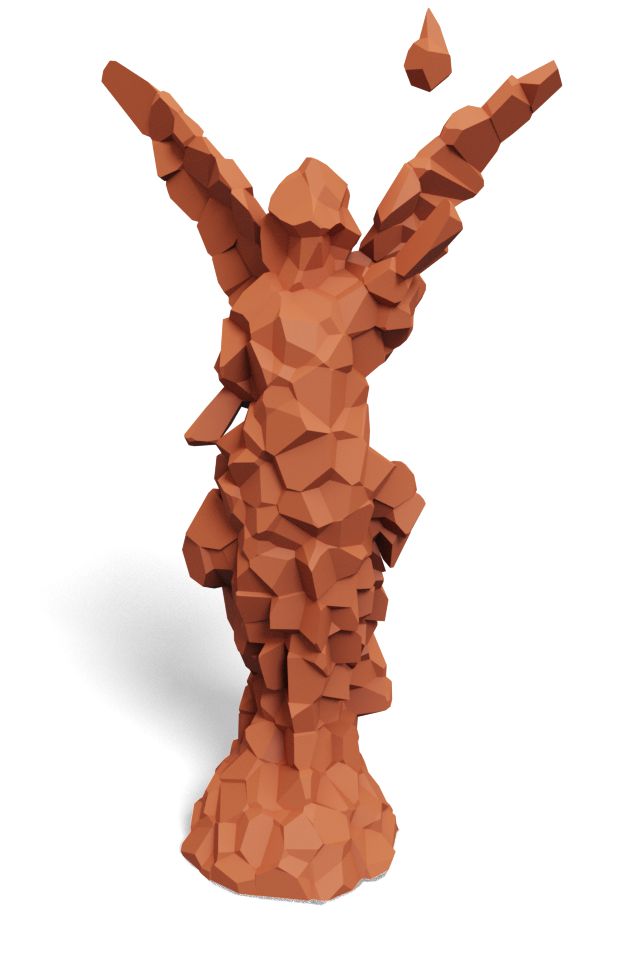}{}
  \colimage{.18}{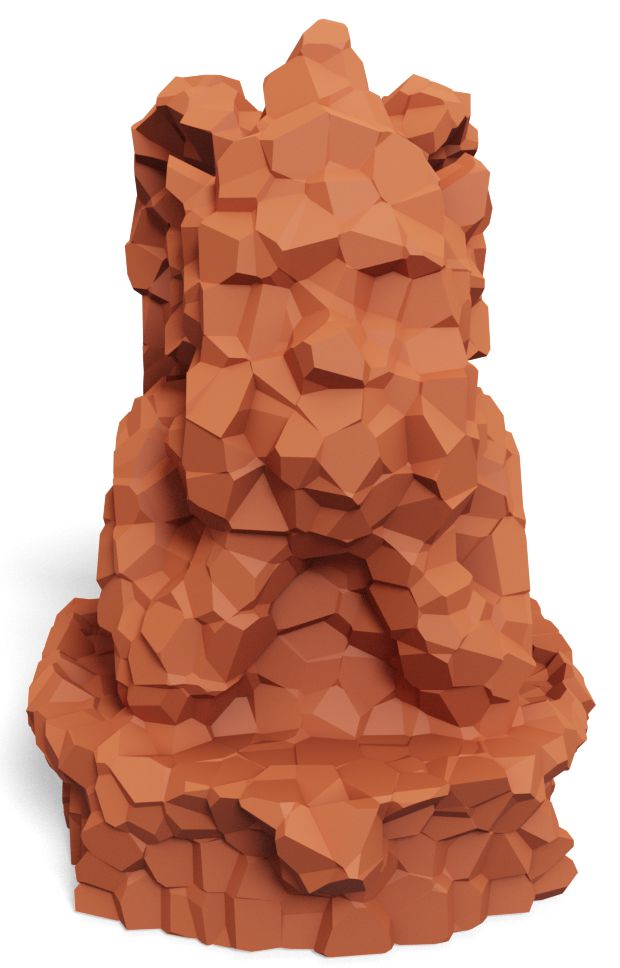}{}
  \colimage{.18}{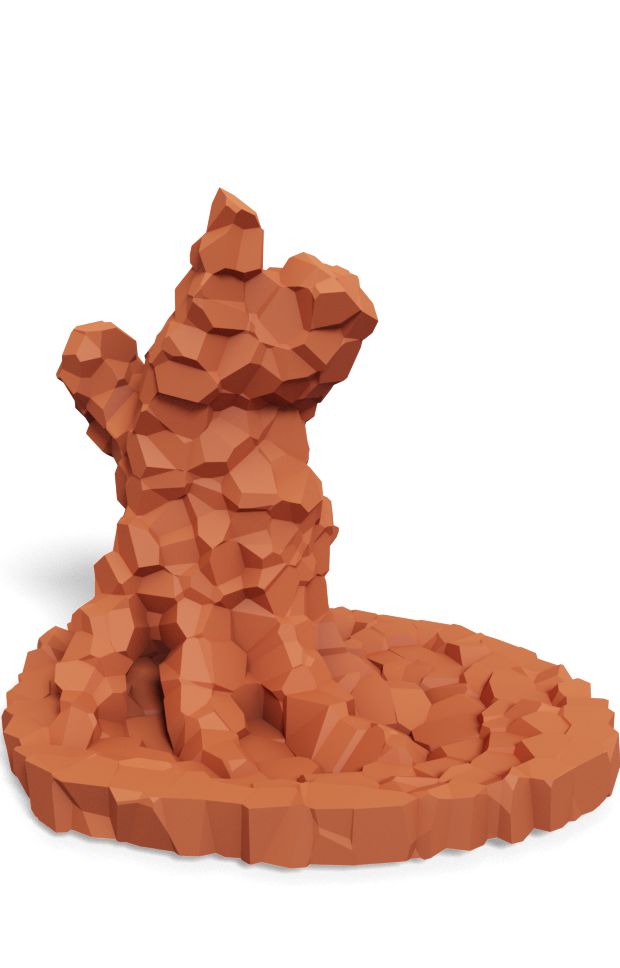}{}
  \vskip 0pt
   \colimage{.18}{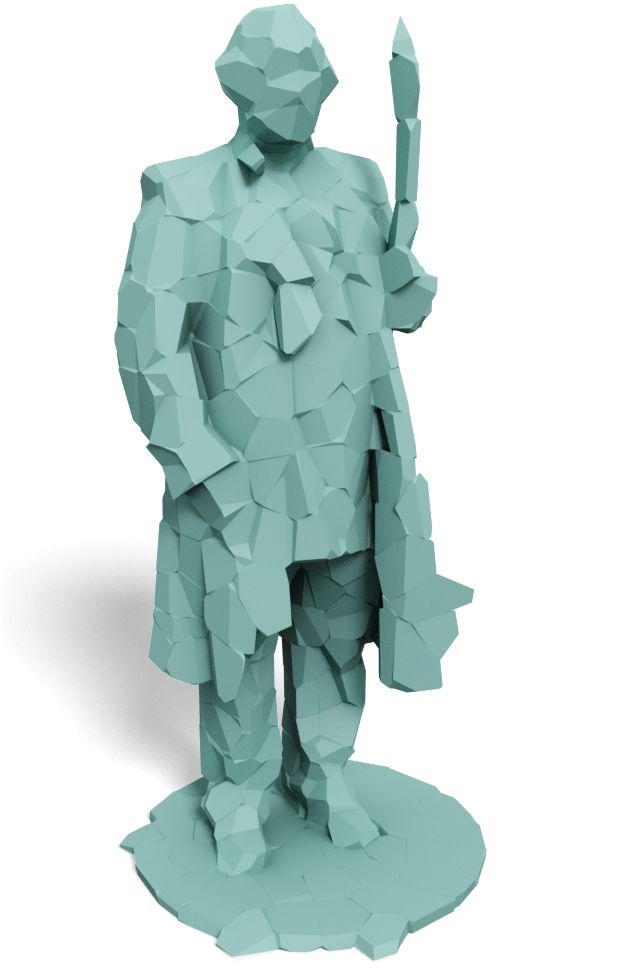}{}
  \colimage{.18}{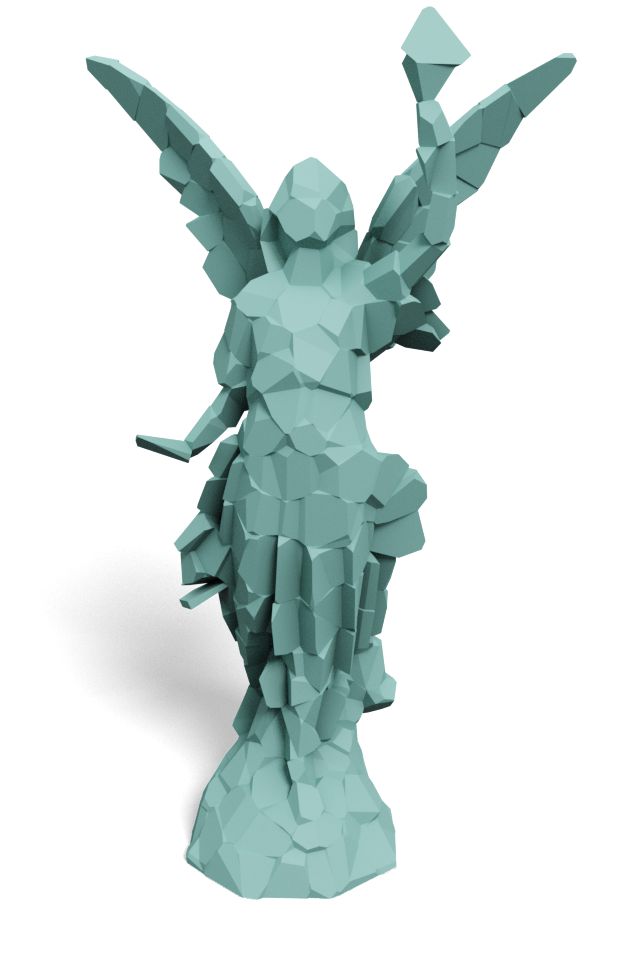}{}
  \colimage{.18}{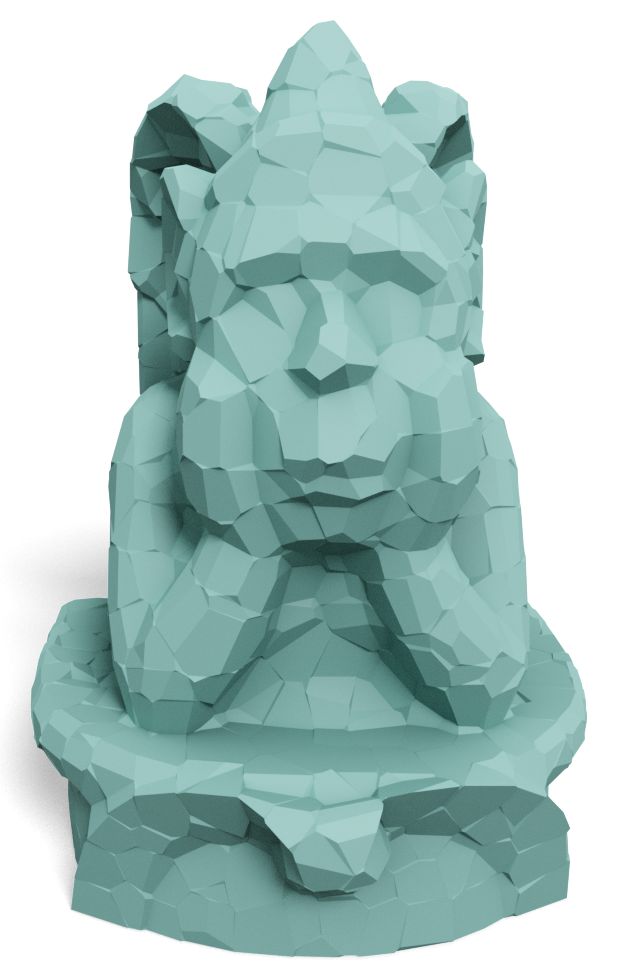}{}
  \colimage{.18}{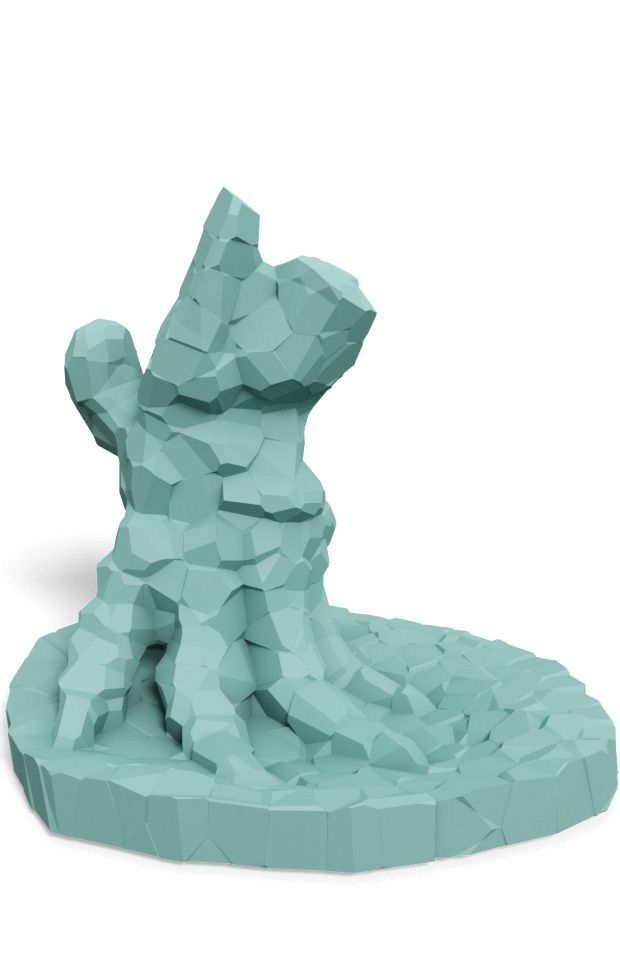}{}
    \vskip 0pt
 \colimage{.18}{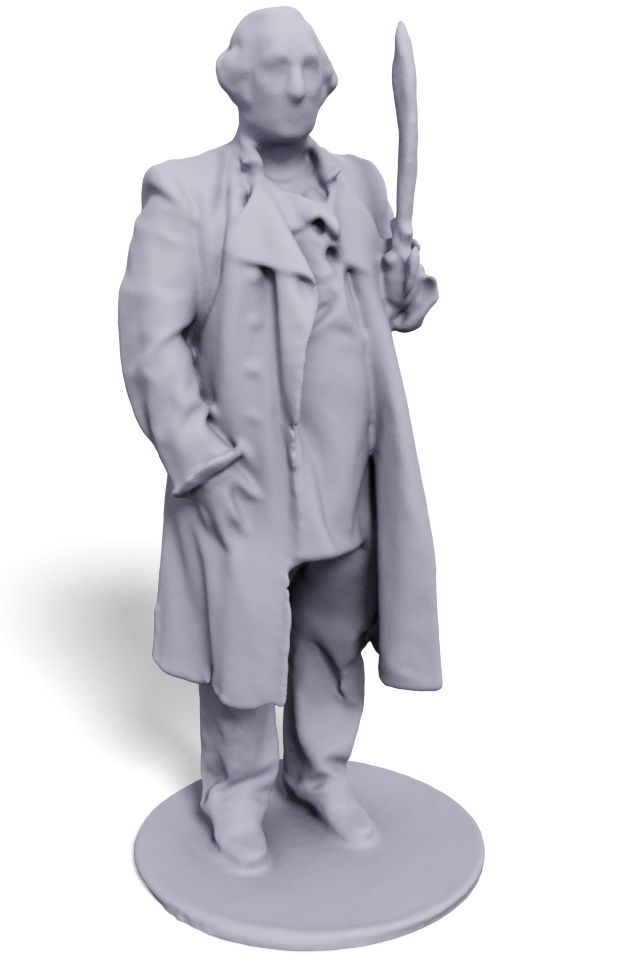}{}
 \colimage{.18}{figs/vorogt/25211gt.jpeg}{}
 \colimage{.18}{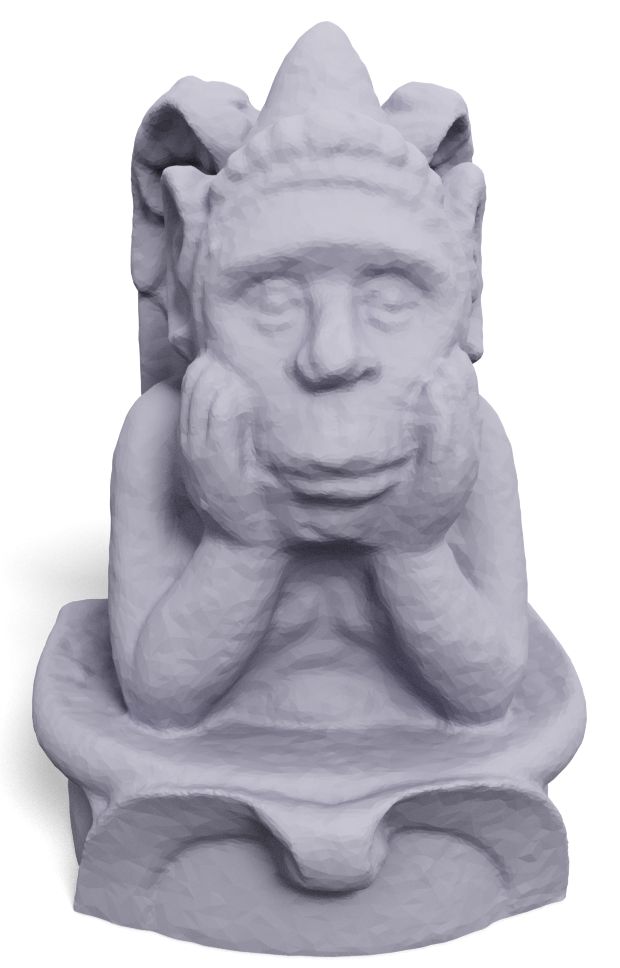}{}
 \colimage{.18}{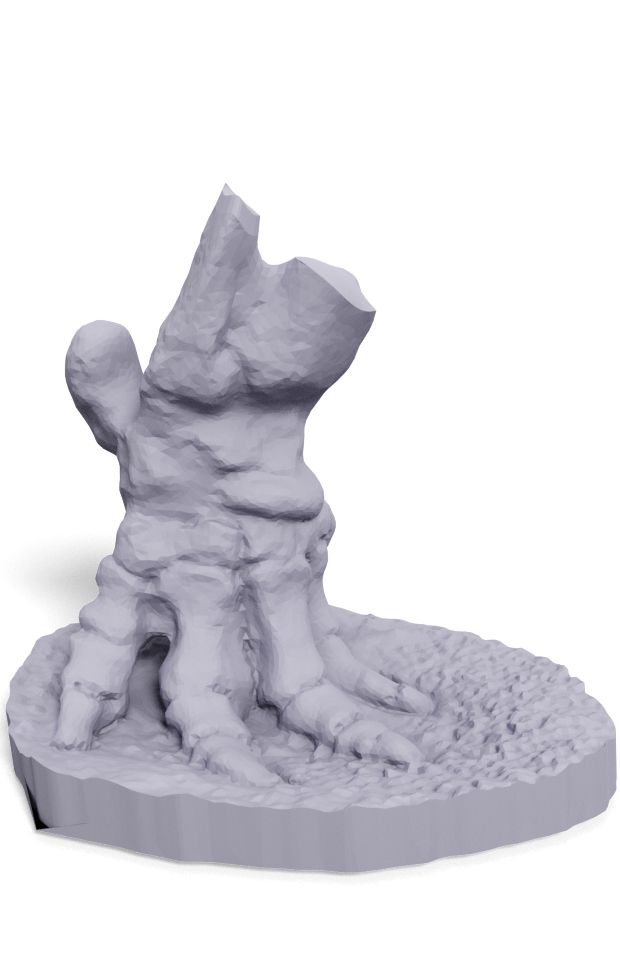}{}
   \caption{VoronoiNet~\cite{williams2020voronoinet} (top row), our method (middle row) and target shape (bottom row) for a grid of size $32^3$}\label{fig:voro}
\end{figure*}

The optimization-based results are presented in Figures \ref{fig:voro}, \ref{fig:dir1}, \ref{fig:dir2}, \ref{fig:dir3}, and \ref{fig:dir4}. Our representation is accurate and efficiently captures small details, visually outperforming all methods. 

The learning-based results on the test set of ABC are presented in Figures \ref{fig:abc1}, \ref{fig:abc2}, and \ref{fig:abc3}, while Thingi32 shapes are in Figures \ref{fig:thingi1}, \ref{fig:thingi2}. Our approach captures thin structures, but sometimes fails to correctly predict the occupancy of some cells, resulting in shadowed voids. 

\begin{figure}[!h]\vspace*{0mm}
 \centering\begin{subfigure}{.22\textwidth}
  \centering
\includegraphics[width=\linewidth]{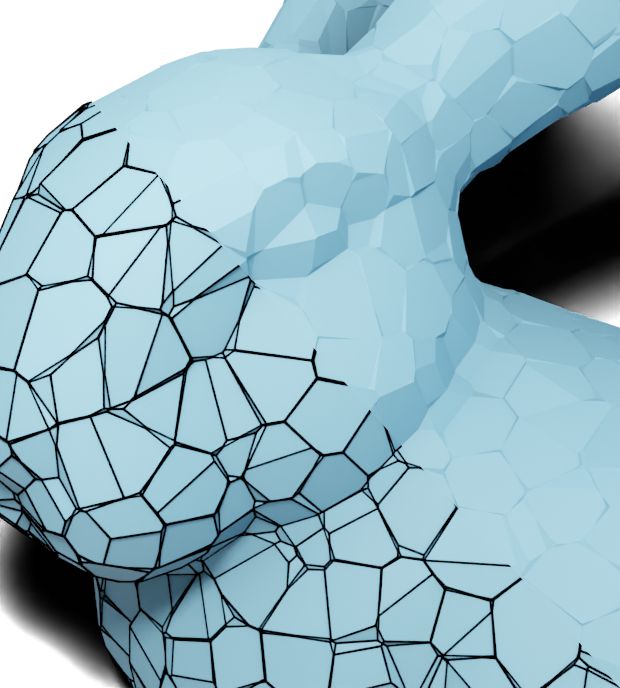}
\caption{Voromesh}\label{fig:artivoro}
\end{subfigure}
\hfill
\begin{subfigure}{.22\textwidth}
  \centering
\includegraphics[width=\linewidth]{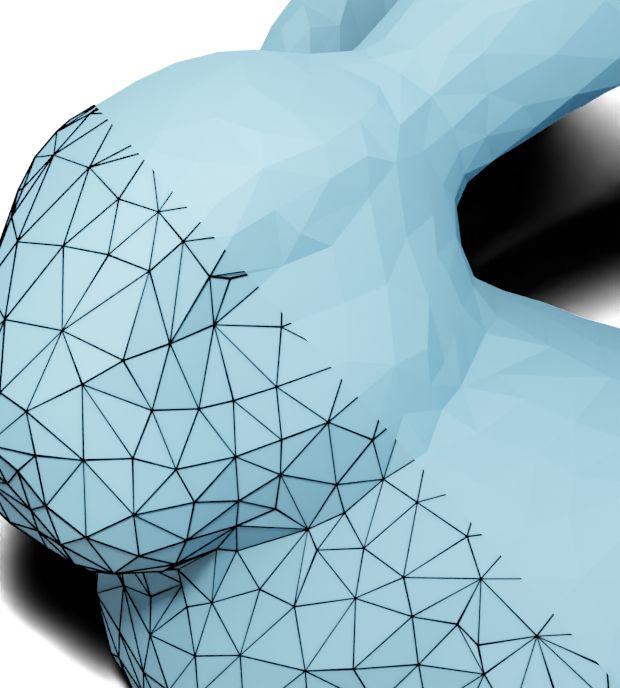}
\caption{Simplified dual Voromesh}\label{fig:artidual}
\end{subfigure}\vspace*{-2mm}
\caption{A possible approach to eliminating artifacts caused by small facets (a) is to dualize the Voromesh, and collapse small edges (b).\vspace*{-4mm}}
\end{figure}

Note finally that the small faces visible on close inspection could be eliminated in a post-processing stage as mentioned as future work in the paper and discussed in the next section --- see also Figures~\ref{fig:artivoro}-\ref{fig:artidual}. 

\section{Surface Artifacts}\label{sec:artifacts}

\begin{figure}[h] \vspace*{-1mm}
  \includegraphics[width=\linewidth]{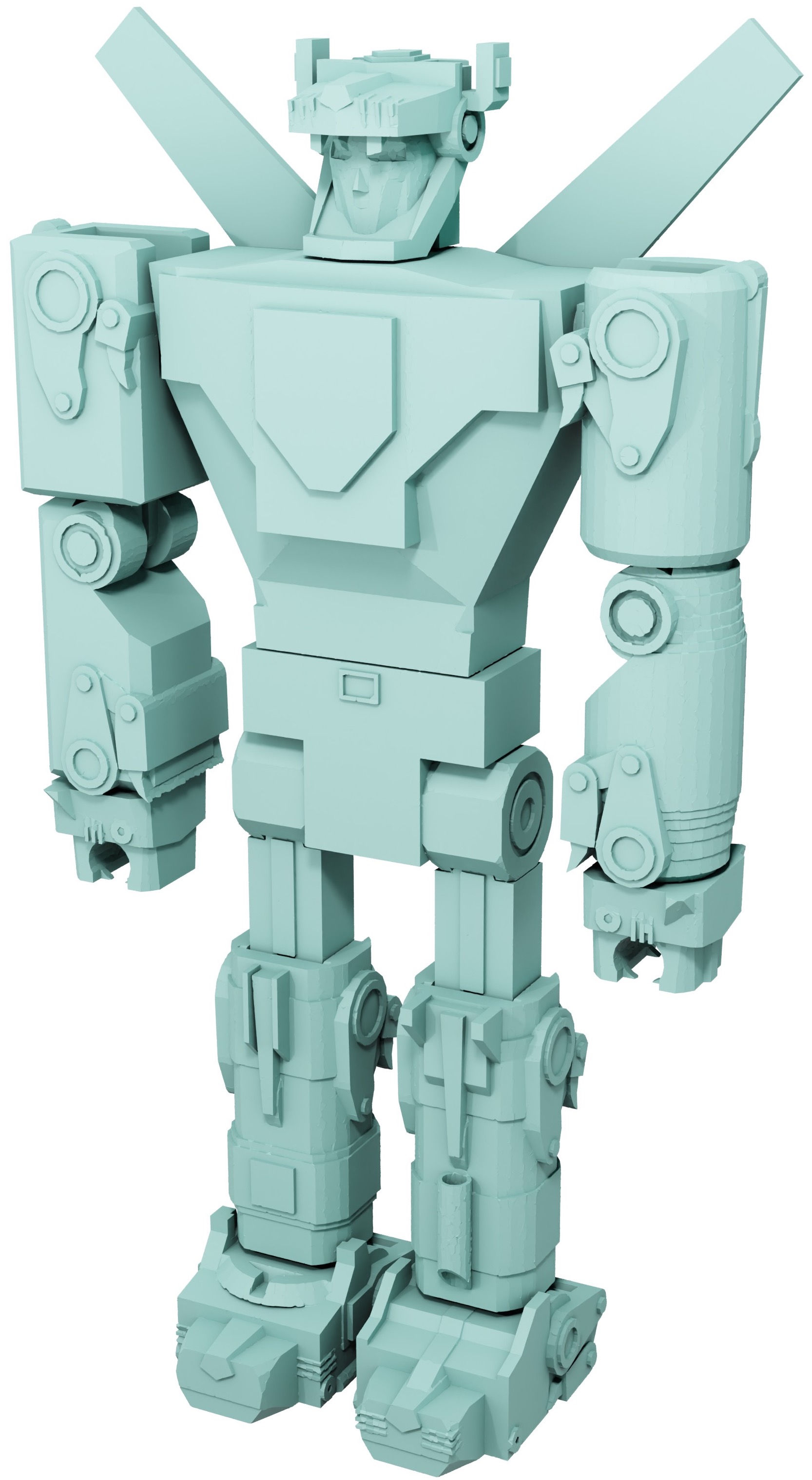}\vspace*{-2mm}
 \caption{Direct optimization on a $256^3$ grid. \vspace*{-3mm}}\label{fig:voltron}
\end{figure}

Voronoi generators in near co-circular positions can create small facets (visible in \fig{artivoro}) which can cause shading artifacts. Removing them would improve visual quality and performance in F1 and NC metrics. In this paper, we favored strong topological guarantees over appearance; nonetheless, we explored ways to improve our reconstructions through postprocessing.
One possibility is to leverage the fact that the \textit{surface dual} to our \textit{VoroMesh} is a triangle mesh, which can be simplified through edge collapses to yield an artifact-free surface with better shaped triangles (see Figure~\ref{fig:artidual}), potentially at the expense of some details.
Alternatively, our method is capable of \textit{very fine} reconstructions where artifacts simply become invisible, see Figure~\ref{fig:voltron}.

\section{Comparison to Deep Marching Tetrahedra}\label{sec:DMTet}

\begin{figure}[htb]
 \centering
 \begin{subfigure}{.11\textwidth}
  \centering
  \includegraphics[width=\linewidth]{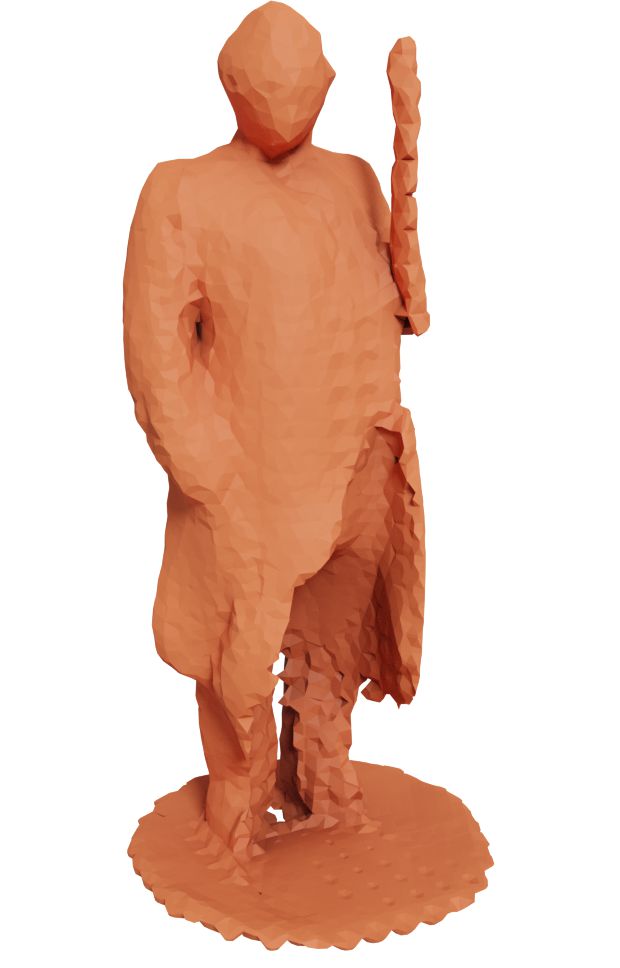}
  \caption{DMT}\label{fig:dmt_tutorial}
\end{subfigure}
\begin{subfigure}{.11\textwidth}
  \centering
  \includegraphics[width=\linewidth]{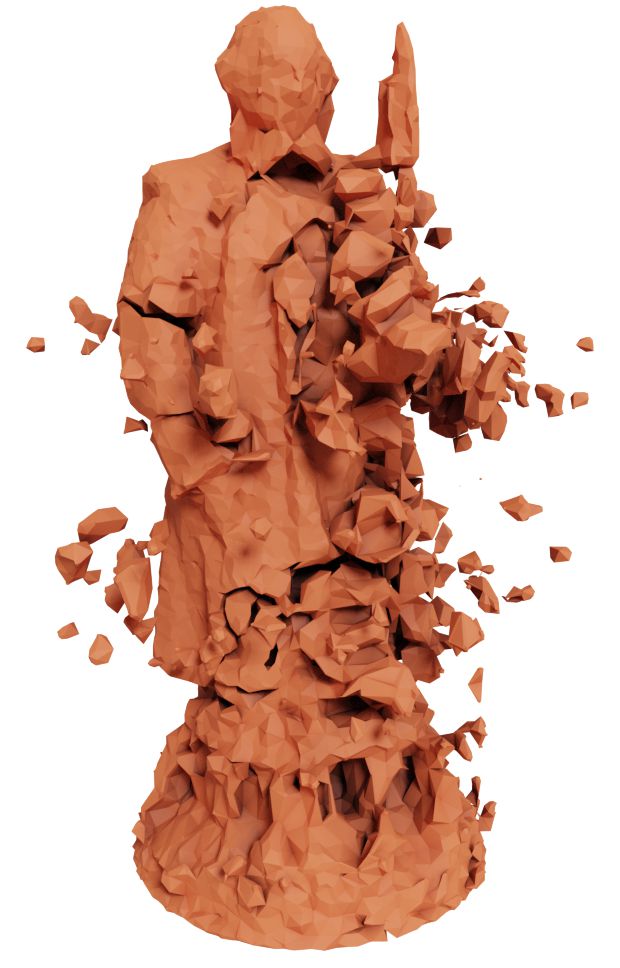}
  \caption{DMT-d1}\label{fig:dmt_direct_sphere}
\end{subfigure}
\begin{subfigure}{.11\textwidth}
  \centering
 \includegraphics[width=\linewidth]{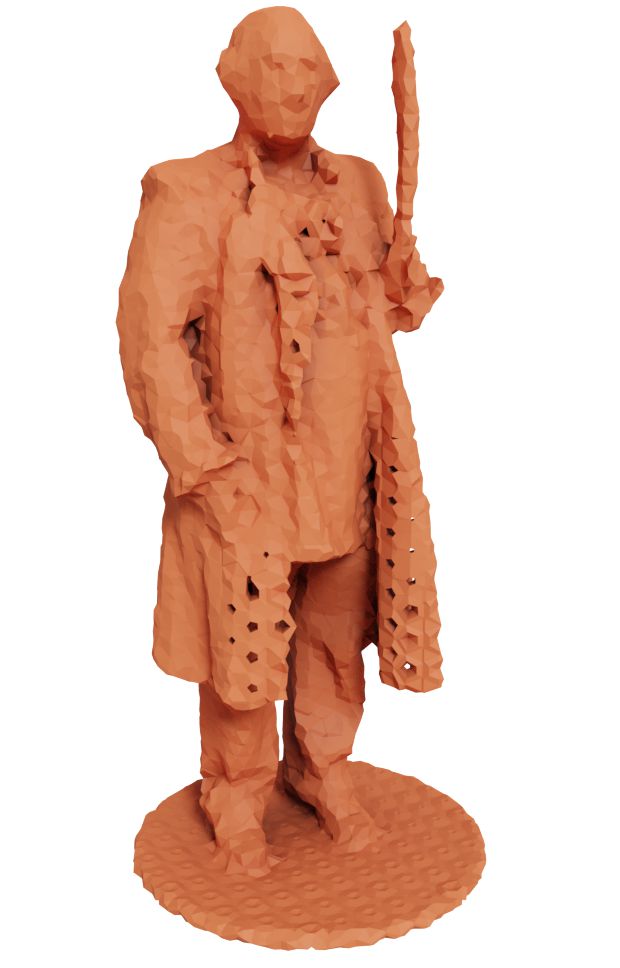}
  \caption{DMT-d2}\label{fig:dmt_direct}
\end{subfigure}
\begin{subfigure}{.11\textwidth}
  \centering
\includegraphics[width=\linewidth]{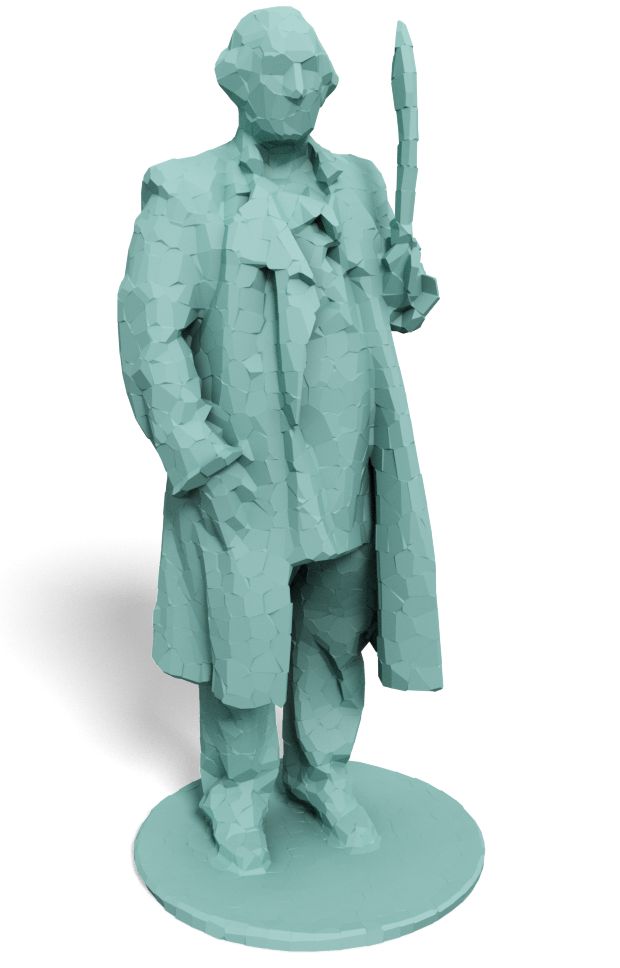}
  \caption{Ours}\label{fig:dmt_ours}
\end{subfigure}
\vspace*{-2mm}
\caption{Visual comparison between DMT (tetrahedra grid of size $128^3$) and our method (grid of size $64^3$). Left to right: vanilla DMT version, DMT with direct optimisation, DMT with direct optimisation and ground-truth SDF initialization, and VoroMesh.\vspace*{2mm}}
\end{figure}

\begin{table}[!h]
  \begin{center}
    \begin{tabular}{|l|c|c|c|}
      \hline
      \textbf{Method} & CD & F1 & NC   \\
      \hline
      DMT (128-d2) & 0.712 & 0.926 & 0.962\\
      VoroMesh (64)  & \textbf{0.645} & \textbf{0.938} & \textbf{0.975}\\
      \hline
      \end{tabular}
    \caption{Quantitative comparisons for an optimization-based 3D reconstruction task on the Thingi32 dataset.\vspace*{-3mm}}
    \label{tab:direct}
  \end{center}
\end{table}

For completeness, we provide additional analysis and comparison of our approach to Deep Marching Tetrahedra~\cite{shen21neurips} (DMT). 
Several critical differences exist between DMT and our Voromesh:
\begin{itemize}
    \item the output connectivity of DMT meshes is fixed by the template, while ours is flexible and determined by construction of a corresponding Voronoi diagram;


    \item the representation of DMT is hybrid as a MLP is necessary to model a continuous implicit field of SDF and displacements, whereas our VoroMesh can be used directly without neural networks.
    \item DMT does not provide strong topological guarantees, namely the absence of self-intersections.
\end{itemize}

In the absence of an official implementation from the DMT paper, we rely on a tutorial code from a separate NVIDIA library\footnote{\href{https://github.com/NVIDIAGameWorks/kaolin/blob/master/examples/tutorial/dmtet_tutorial.ipynb}{DMT tutorial}.}, which overfits a single MLP to a given shape --- a setting similar to our direct optimisation experiment. It relies on a tetrahedra grid of size 128, which has roughly the same number of vertices as a voxel grid of resolution 65 (we used 64 in our experiment). Using it out-of-the-box yields poor results, see Figure~\ref{fig:dmt_tutorial}. We believe it can be attributed to the underlying hybrid model; more precisely, to the inability of the MLP to represent high-frequency details as explained in~\cite{sitzmann2020implicit}. We also tried to optimize the tetrahedra vertices displacement and signed distance predictions directly; because the parameters are now independent and no longer predicted by an MLP, the initial unit sphere fitting leads to worse results, see Figure~\ref{fig:dmt_direct_sphere}: the loss function is unable to optimize the tetrahedra situated far from the target surface. To alleviate this problem, we further help DMT by initializing the displacements to zero and the predicted signed distance to the ground-truth SDF with respect to the original surface; but resulting reconstructions still lack surface smoothness and finer details and are outperformed by our VoroMesh representation, see Figure~\ref{fig:dmt_direct} and Table~\ref{tab:direct}.

\section{Additional timings}\label{sec:timings}

We now provide timings for optimization-based and learning-based experiments. Our mesh extraction, which relies on CGAL, is very fast (\tab{optimtimings}), while our inference time for the learning-based experiment is comparable to state-of-the-art methods (\tab{learningtimings}).

\begin{table}[h]
  \begin{center}
        \begin{tabular}{|l|c||c|}
          \hline
            Grid Size & Mesh Extraction (s) & Full Execution (s)  \\
          \hline    
        32 & 0.03 & 5.0  \\
        64 & 0.11 & 10.2 \\
        128 & 0.46 & 57.3 \\
        \hline
        \end{tabular}%
    \caption{Mean timings on the Thingi32 dataset for optimization-based 3D reconstruction. Full execution timings include mesh extraction.}
    \label{tab:optimtimings}
  \end{center}
\end{table}

\begin{table}[h]
  \begin{center}
        \begin{tabular}{|l|c|c|}
          \hline
          Method & Grid Size& Mean Inference (s) \\
          \hline    
        NDC~\cite{chen22tog} & 32  & 0.05 \\
        NMC~\cite{chen21tog} & 32  & 0.18 \\
        Ours & 32  & 0.19\\
          \hline    
        NDC~\cite{chen22tog} & 64  & 0.12 \\
        NMC~\cite{chen21tog} & 64  & 0.97 \\
        Ours & 64 & 0.63\\
        \hline
        \end{tabular}%
    \caption{Mean timings on the ABC test set for learning-based 3D reconstruction.}
    \label{tab:learningtimings}
  \end{center}
\end{table}

\color{black}
 
\pageimage{.18}{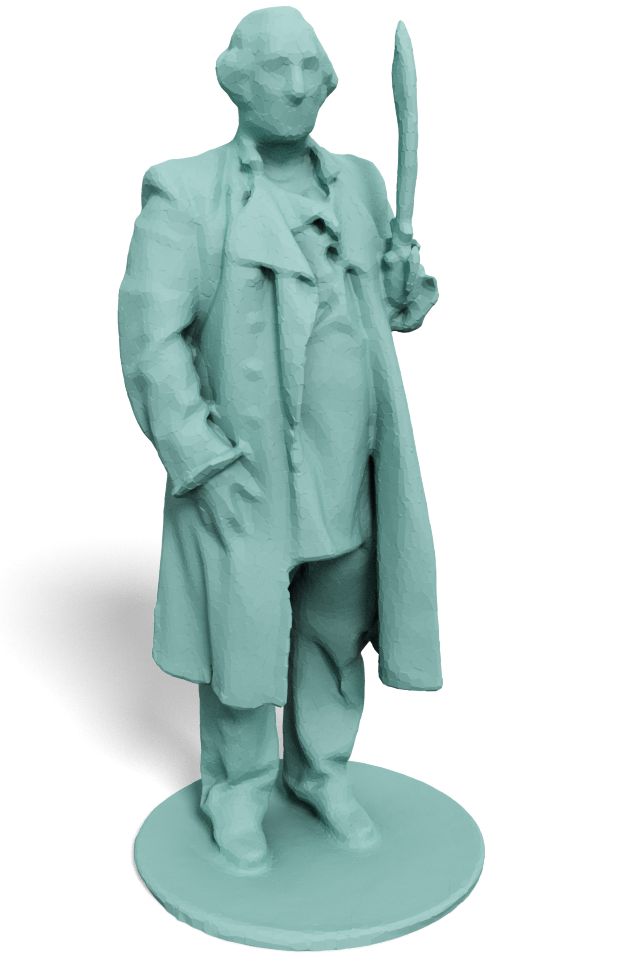}{fig:dir1}
\pageimage{.18}{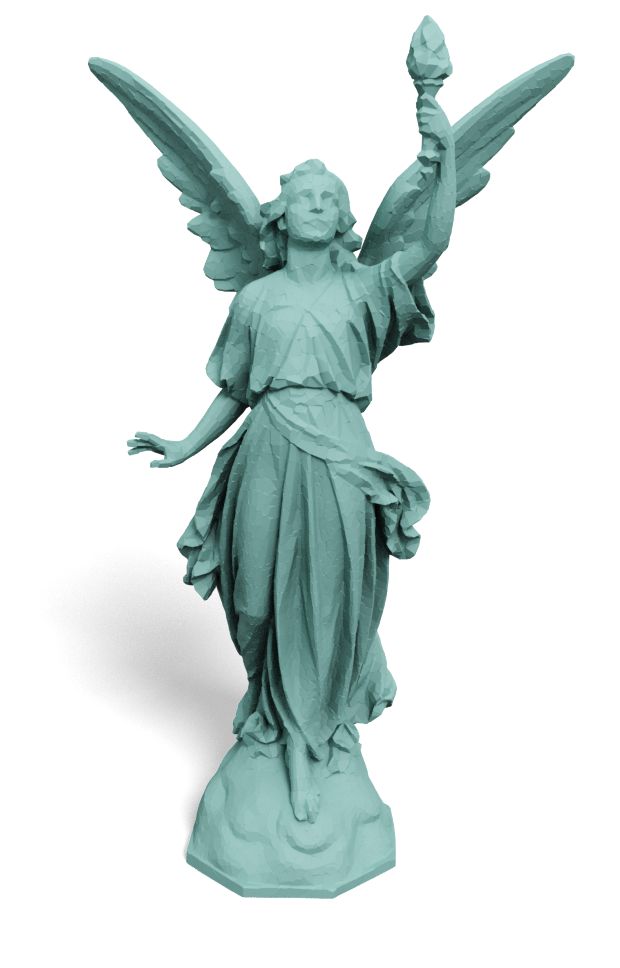}{fig:dir2}
\pageimage{.18}{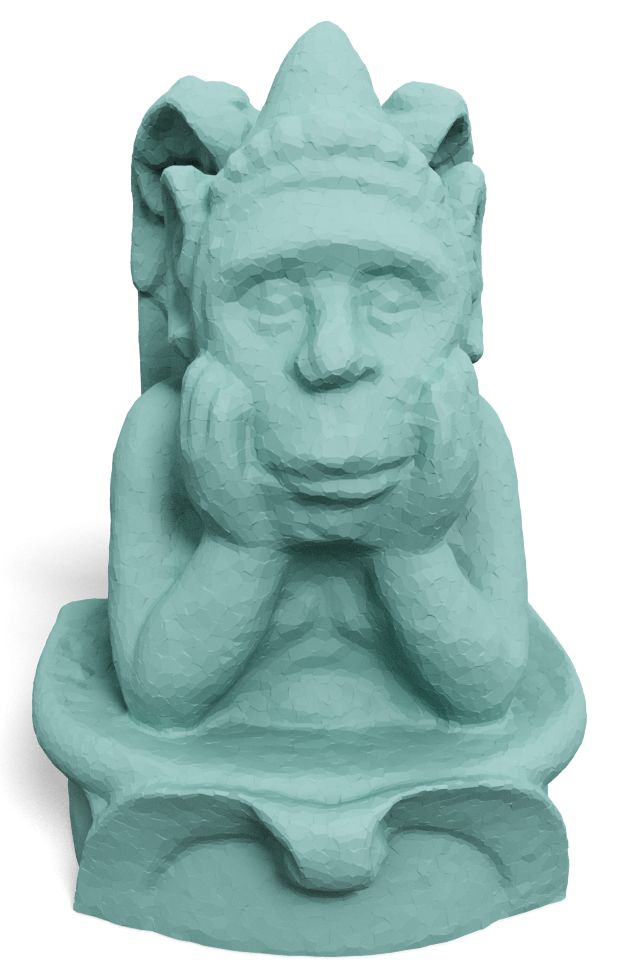}{fig:dir3}
\pageimage{.18}{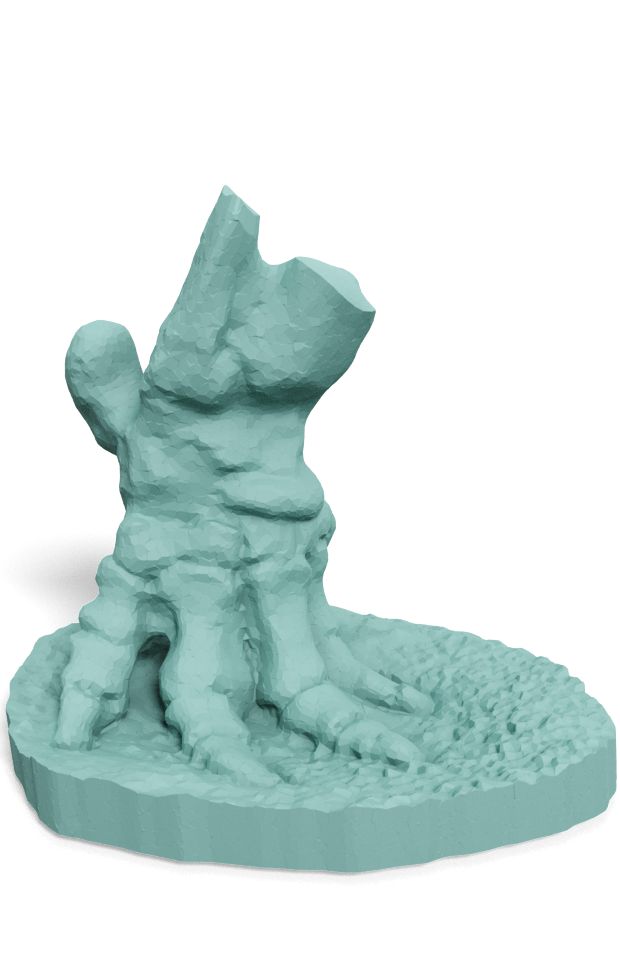}{fig:dir4}

\abcpageimage{.23}{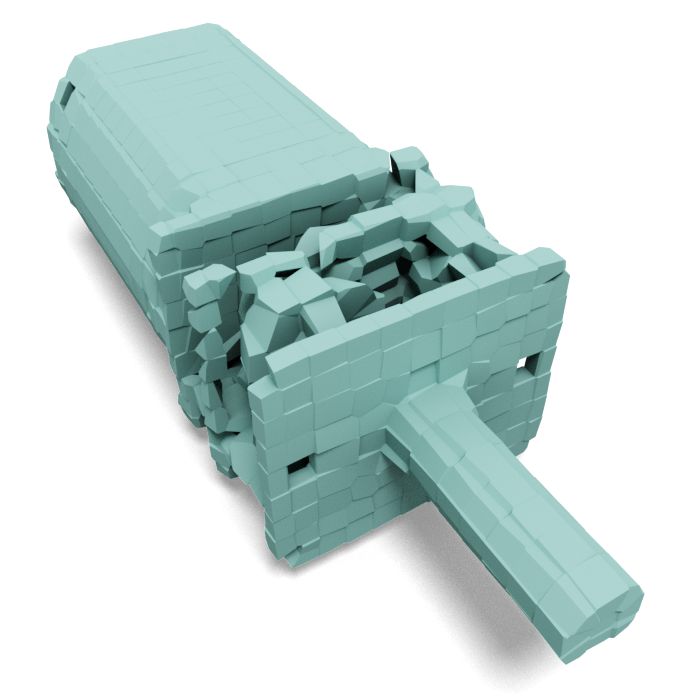}{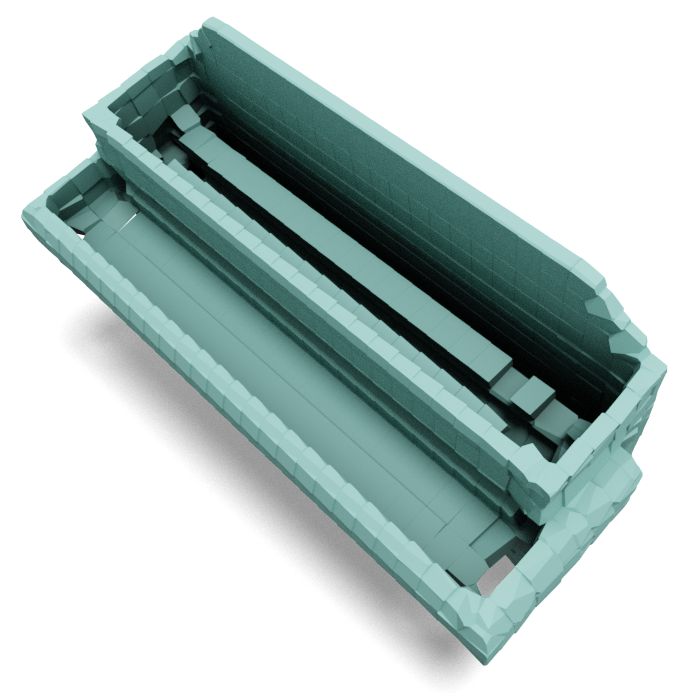}{fig:abc1}
\abcpageimage{.23}{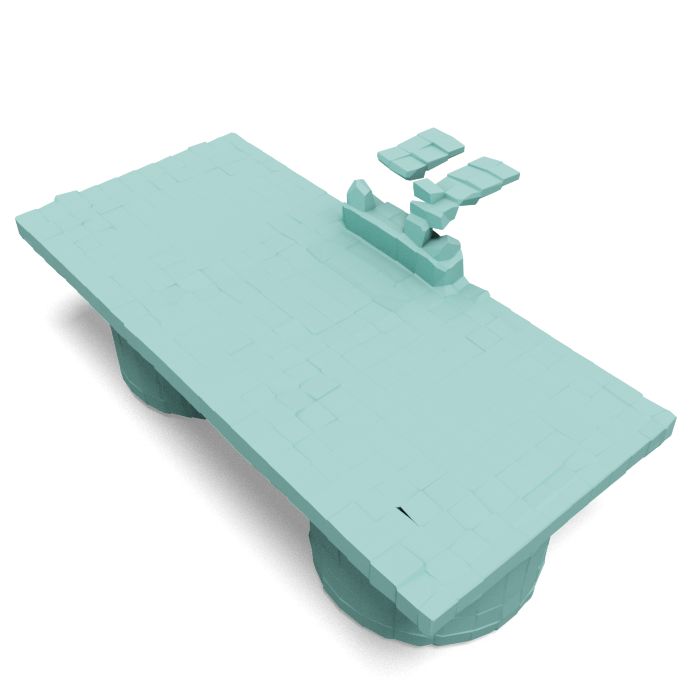}{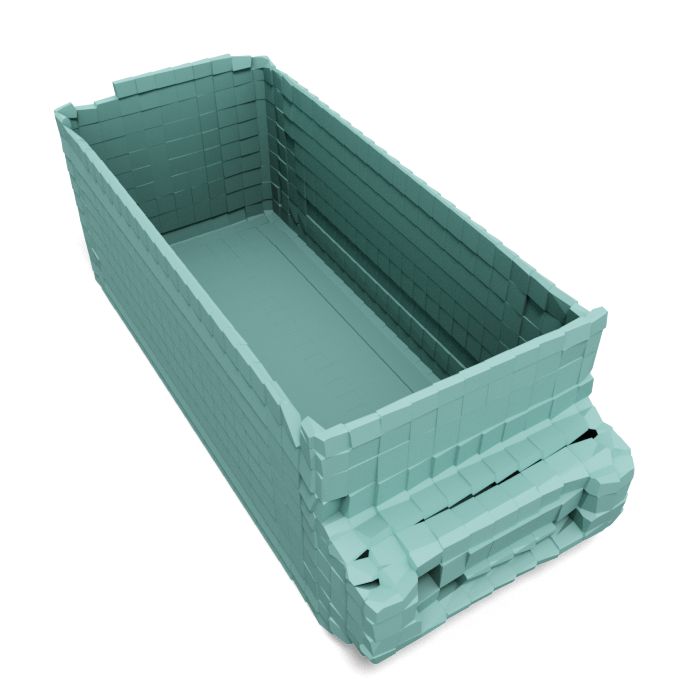}{fig:abc2}
\abcpageimage{.23}{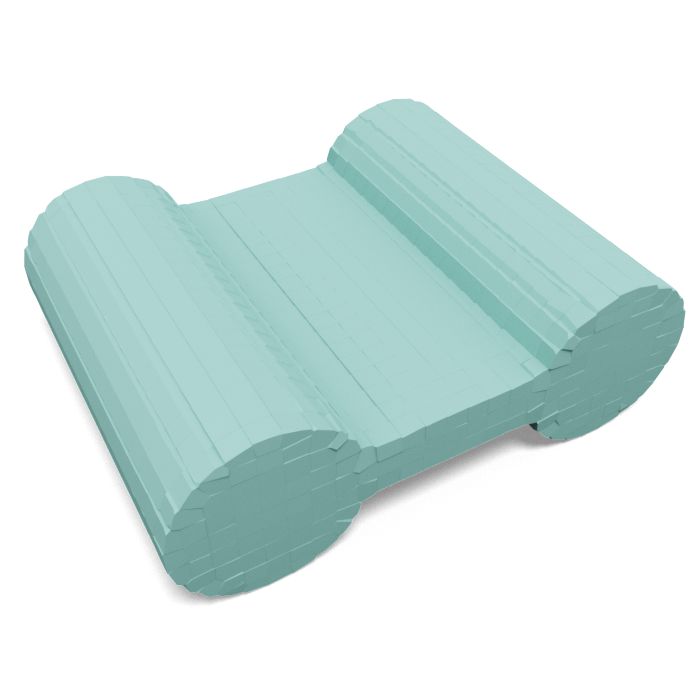}{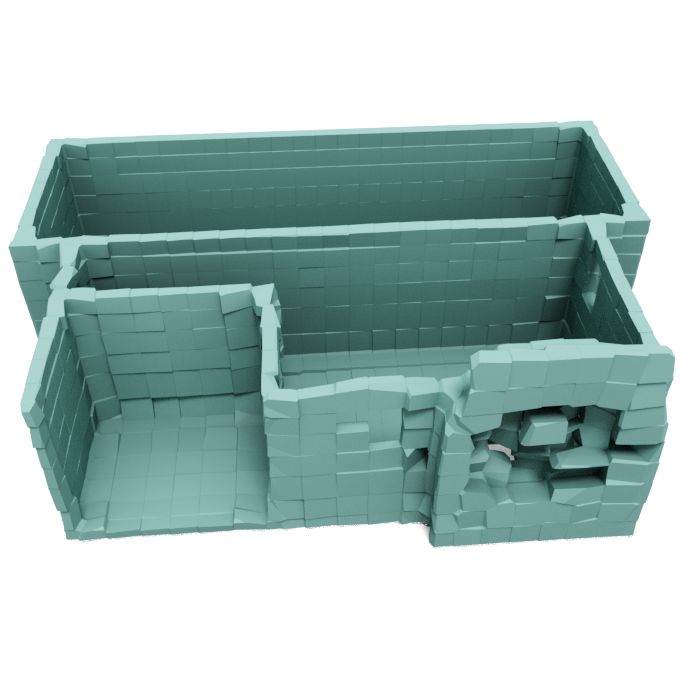}{fig:abc3}

\thingipageimage{.23}{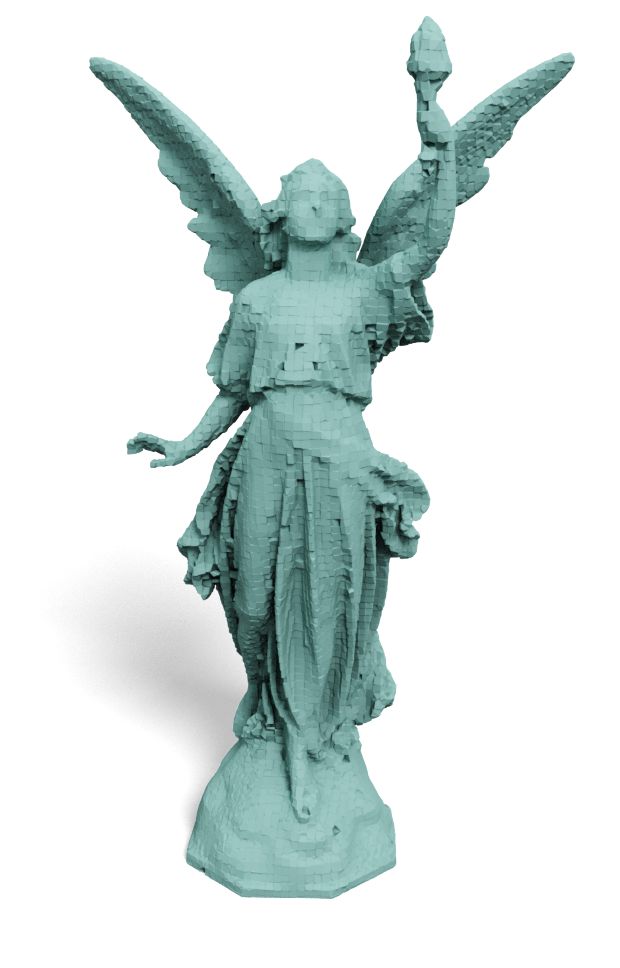}{fig:thingi1}
\thingipageimage{.23}{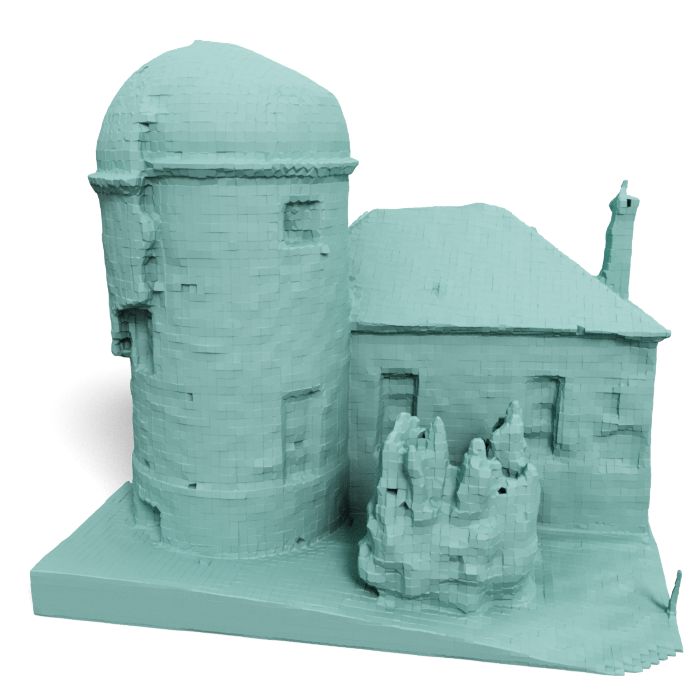}{fig:thingi2}

\end{document}